\documentclass[twoside]{article}

\usepackage[accepted]{aistats2021}

\usepackage[round]{natbib}

\usepackage[utf8]{inputenc} %
\usepackage[T1]{fontenc}    %
\usepackage{hyperref}       %
\usepackage{url}            %
\usepackage{booktabs}       %
\usepackage{amsfonts}       %
\usepackage{nicefrac}       %
\usepackage{microtype}      %

\usepackage{microtype}
\usepackage{graphicx}
\usepackage[algoruled]{algorithm2e}
\SetCommentSty{mycommfont}
\usepackage{algorithmic}

\usepackage{natbib}

\usepackage{xcolor} %
\colorlet{darkblue}{blue!50!black} %
\hypersetup{colorlinks=true}
\hypersetup{citecolor=darkblue} %
\hypersetup{linkcolor=darkblue} %
\hypersetup{urlcolor=darkblue} %

\usepackage{enumitem} 
\usepackage{amsmath,amssymb,amsthm}
\usepackage{mathtools}
\usepackage{subcaption}
\usepackage{dsfont}
\usepackage{wrapfig}
\newcommand{\mathbold}[1]{\ensuremath{\boldsymbol{\mathbf{#1}}}}

\DeclareRobustCommand{\KL}[2]{\ensuremath{\textrm{KL}\left(#1\;\|\;#2\right)}}
\DeclareRobustCommand{\Ep}[2]{\ensuremath{\mathds{E}_{#1}\left[#2\right]}}

\newcommand{\Norm}{\mathcal{N}}

\newcommand{\veps}{\varepsilon}

\DeclareMathOperator*{\argmax}{arg\,max}

\newcommand{\mbh}{\mathbold{h}}

\newcommand{\mbt}{\mathbold{t}}

\newcommand{\mby}{\mathbold{y}}

\newcommand{\mbH}{\mathbold{H}}

\newcommand{\mbJ}{\mathbold{J}}

\newcommand{\mbY}{\mathbold{Y}}

\newcommand{\mbgamma}{\mathbold{\gamma}}

\newcommand{\mbrho}{\mathbold{\rho}}

\newcommand{\mbtheta}{\mathbold{\theta}}

\newcommand{\mbSigma}{\mathbold{\Sigma}}

\newcommand{\mcC}{\mathcal{C}}
\newcommand{\mcD}{\mathcal{D}}

\newcommand{\mcN}{\mathcal{N}}

\newcommand{\mdE}{\mathds{E}}

\newcommand{\mdP}{\mathds{P}}

\newcommand{\mdR}{\mathds{R}}

\newcommand{\tf}{{\theta_f}}

\usepackage{tikz}
\usetikzlibrary{shapes.geometric, arrows, fit, automata, positioning, calc, arrows.meta}
\usetikzlibrary{backgrounds}
\tikzstyle{box} = [rectangle, rounded corners, minimum width=3cm, minimum height=1cm,text centered, draw=black, inner sep=7pt]
\tikzset{
  plate/.style={draw, shape=rectangle, rounded corners=0.5ex, thick,
    minimum width=3.1cm, text width=3.1cm, align=right, inner sep=10pt, inner ysep=10pt, 
    append after command={node[above left= 3pt of \tikzlastnode.south east] {#1}}}
}
\usepackage{adjustbox}
\usepackage{cancel}
\usepackage{multirow}

\usepackage{xargs}
\usepackage[colorinlistoftodos,disable]{todonotes}

\newcommandx{\rtodo}[2][1=]{\todo[linecolor=red,backgroundcolor=red!25,bordercolor=red,inline,#1]{Rebuttal-todo: #2}}
\newcommandx{\rdone}[2][1=]{\todo[linecolor=red,backgroundcolor=green!25,bordercolor=red,inline,#1]{todo was: #2}}

\allowdisplaybreaks %
\newcommand\blfootnote[1]{%
  \begingroup
  \renewcommand\thefootnote{}\footnote{#1}%
  \addtocounter{footnote}{-1}%
  \endgroup
}

\begin{document}

\runningauthor{Hau\ss mann, Gerwinn, Look, Rakitsch, Kandemir}
\runningtitle{Learning Partially Known Stochastic Dynamics with Empirical PAC Bayes}
\twocolumn[

\aistatstitle{Learning Partially Known Stochastic Dynamics\\ with Empirical PAC Bayes}

\aistatsauthor{Manuel Hau\ss mann$^{1*}$ \And Sebastian Gerwinn$^{2*}$ \And Andreas Look$^{2}$}
\aistatsauthor{Barbara Rakitsch$^{2}$ \And Melih Kandemir$^{2}$}

\aistatsaddress{${}^{1}$HCI/IWR, Heidelberg University\\ \texttt{manuel.haussmann@iwr.uni-heidelberg.de}\\ Heidelberg, Germany \And ${}^2$Bosch Center for Artificial Intelligence\\ \texttt{firstname.lastname@de.bosch.com}\\ Renningen, Germany} ]

\begin{abstract}
Neural Stochastic Differential Equations model a dynamical environment with neural nets assigned to their drift and diffusion terms. The high expressive power of their nonlinearity comes at the expense of instability in the identification of the large set of free parameters. This paper presents a recipe to improve the prediction accuracy of such models in three steps: i) accounting for epistemic uncertainty by assuming probabilistic weights, ii) incorporation of partial knowledge on the state dynamics, and iii) training the resultant hybrid model by an objective derived from a PAC-Bayesian generalization bound. We observe in our experiments that this recipe effectively translates partial and noisy prior knowledge into an improved model fit.
\end{abstract}

\section{INTRODUCTION}
\label{sec:intro}

In many engineering applications, it is often easy to model dominant characteristics of a dynamical environment by a system of differential equations with a small set of state variables. In contrast, black-box machine learning methods are often highly accurate but less interpretable. Pushing the model towards high fidelity by capturing intricate properties of the environment, however, usually requires highly flexible, e.g.\ over-parameterized models. Fitting these models to data can, in turn, result in over-fitting and hence poor generalization ability due to their high capacity. 

Our work combines the benefits of both types of models by hybrid modeling: We set up the learning task as a non-linear system identification problem with partially known system characteristics. It assumes to have access to a differential equation system that describes the dynamics of the target environment with low fidelity, e.g.\ by describing the vector field on a reduced dimensionality, by ignoring detailed models of some system components, or by avoiding certain dependencies for computational feasibility. We incorporate the ODE system provided by the domain expert into a non-linear system identification engine, which we choose to be a \emph{Bayesian Neural Stochastic Differential Equation}~(BNSDE) to cover a large scope of dynamical systems, resulting in a \emph{hybrid model}.\blfootnote{${}^*$ Equal contribution.}

We propose a new algorithm for stable and effective training of such a hybrid BNSDE that combines the strengths of two statistical approaches: i) Bayesian model selection \citep{williams2006gaussian}, and ii) Probably Approximately Correct~(PAC) Bayesian bounds \citep{macallester1999pac, seeger2002pac}. 
We improve the theoretical links between these two approaches \citep{germain2016pac} by demonstrating how they can co-operate \emph{during} training. To this end, we propose a novel training objective that suits SDE inference and derive a PAC-Bayesian generalization bound. Further, we provide a proof that this bound is upper bounded by the marginal likelihood of the BNSDE hyperparameters and a complexity penalizer. Gradients of this upper bound are {\it tied} to the actual PAC bound, hence tightening the upper bound also tightens the PAC bound. Consequently, optimizing this bound amounts to Empirical Bayes stabilized by a regularizer developed from first principles. We refer to using this objective for training as {\it Empirical PAC-Bayes}.

We demonstrate that our method can translate coarse descriptions of the actual underlying dynamics into a consistent forecasting accuracy increase. We first show the necessity of each of the multiple steps that comprise our method in an ablation study. Finally, we demonstrate in a real-world motion capture modelling task, that our method outperforms black-box system identification approaches \citep{chen2018neural, hegde19differential, look2019diff} and alternative hybridization schemes that incorporate second-order Newtonian mechanics \citep{yildiz19odevae}.

\section{BACKGROUND}
Our contribution combines approaches from stochastic differential equations, PAC-Bayes, and Empirical Bayes. Hence, we first introduce each of these concepts.

\paragraph{Stochastic Differential Equations.}\label{sec:background:sde}
Stochastic differential equations (SDEs) are an extension of ordinary differential equations (ODEs) to include stochastic fluctuations in the dynamics \citep{oksendal92stochastic}. If we let ${\bf{h}}_t \in \mdR^P $ denote the $P$-dimensional state, the dynamics can be  written in the following form: 
\begin{align}
    d{\bf{h}}_t & = f({\bf{h}}_t,t) dt + G({\bf{h}}_t,t) dW_t,
\label{eq:SDE}
\end{align}
where the drift term is given by an arbitrary non-linear function $f(\cdot,\cdot): \mdR^P\times \mdR_+ \rightarrow \mdR^P$ and the matrix valued function $G(\cdot,\cdot) :\mdR^P\times \mdR_+ \to \mdR^{P\times P}$ governs the diffusion dynamics. Finally, $W_t$ denotes a $P$-dimensional Wiener Process determining the stochastic fluctuations. The solution to the SDE is a stochastic process ${\bf{h}}_t$. %

As analytical solutions of SDEs are not available except for specific choices of $f$ and $G$, one has to resort to numerical approximation methods. Analogous to the practice for ODEs, a common approach which we follow is to use the Euler-Maruyama (EM) method \citep{sarkka2019applied}, which discretizes the SDE in time steps $t_1,\dots, t_K$, resulting in the following sample-based approximation to the joint distribution:
 \begin{align}
 \begin{split}
 \label{eq:em-discretization}
     {\bf{h}}_{t_{k+1}} & = {\bf{h}}_{t_k} + f({\bf{h}}_{t_k}, t_k) \Delta t_k + G({\bf{h}}_{t_k}, t_k)  \Delta W_k,\\
     \Delta &W_k \sim \mcN(0, \Delta t_k \mathds{1}_P),\quad      \Delta t_k  := t_{k+1} - t_k,
       \end{split}
 \end{align}
 where $\mathds{1}_P$ is a $P$ dimensional identity matrix. 
 Using this sampling scheme, we obtain an approximation to the joint distribution $p({\bf{h}}_{t_1}, \dots {\bf{h}}_{t_K})$ for the given (fixed) drift and diffusion functions. %

\paragraph{PAC-Bayes.} Probably approximately correct (PAC) bounds quantify a model's generalization capabilities from a training set to the true data distribution. To this end, a risk $R(h)=\Ep{x}{l(x,h(x))}$ of a hypothesis $h$  is defined via a loss function $l(x, h(x))$ that measures the loss of the hypothesis evaluated at a data point $x$. Particularly, we build upon the PAC-Bayesian formulation~\citep{macallester1999pac, macallester2003pac}, in which the generalization performance of a posterior, i.e.\ a distribution $Q$ over hypotheses, is characterized by the following bound which holds with probability greater than $1-\delta$:
\begin{equation*}
  \forall Q: \quad \Ep{{h}\sim Q}{R({h})} \le \Ep{{h}\sim Q}{R_\mcD({h})} + \mcC(P,Q, \delta, N).
\end{equation*}
In the inequality above, $\Ep{Q}{R(h)}$ is the expected risk across all hypotheses under the true data distribution, which is not accessible in practice, and $\Ep{Q}{R_\mcD(h)}=\Ep{Q}{\frac{1}{|\mcD|}\sum_{x\in \mcD}{l\big(x,h(x)\big)}}$ is its empirical counterpart in which the risk is averaged across the observed data $\mcD$. A distribution $P$ over the hypotheses referred to as the prior determines the complexity term $\mcC(P,Q, \delta, N)$. This term additionally depends on the number of observed data points~$N$ and a confidence variable~$\delta$ specifying the probability with which the bound holds~\citep{macallester1999pac, maurer2004note}.

\paragraph{Empirical Bayes.}
Bayesian models define a prior distribution $p_\phi(\theta)$ over parameters $\theta$ with hyperparameter $\phi$, which together with the likelihood $p(\mcD|\theta)$ defines the full model. The standard approach consists of learning a posterior over these parameters $p(\theta|\mcD)$ keeping the hyperparameters $\phi$ fixed and marginalizing over $\theta$ in a second step to get the posterior predictive. An alternative, known as Empirical Bayes or Type-II maximum likelihood \citep{bishop2006pattern}, directly marginalizes over the prior, and optimizes the resulting marginal likelihood with respect to the hyperparameters $\phi$,
\begin{equation}
    \phi^* = \argmax_\phi \int p(\mcD|\theta)p_\phi(\theta)d\theta.
\end{equation}

\section{THE PROPOSED METHOD}
In this section, we describe how to combine these tools into a coherent whole for effective inference. We first construct a BNSDE and equip it with domain-specific prior knowledge. Then, we derive a PAC-Bayesian objective to fit it to data and conclude with results on the proposed approach's convergence.

    \pgfdeclarelayer{background}
\pgfdeclarelayer{foreground}
\pgfsetlayers{background,main,foreground}
\begin{figure}
  \input{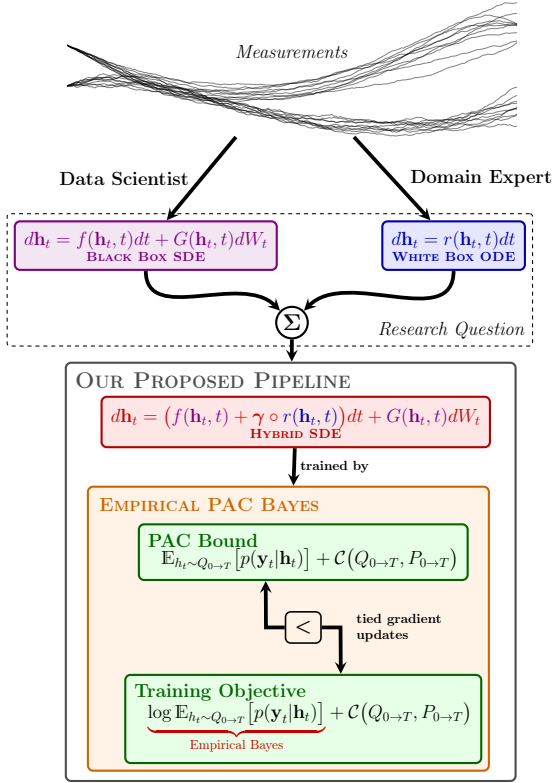}%
  \caption{Illustration of the research question we pose (above) and our proposed solution (below).}
  \label{fig:figureone}
  \end{figure}

\subsection{A Hybrid BNSDE}\label{sec:sdevariations}
Application of deep learning to differential equation modelling paves the way to high-capacity predictors for capturing complex dynamics \citep{chen2018neural, rackauckas2020universal}. Neural Stochastic Differential Equations (NSDEs) \citep{look2019diff, raginsky2019neural} are SDEs as defined in~\eqref{eq:SDE} where the drift function, and potentially also the diffusion function are modelled as neural nets. 
As an initial step towards effective training, we introduce a prior distribution $p_\phi(\tf)$, parameterized by $\phi$ on the weights $\theta_f$ of an NSDE drift network, and arrive at
\begin{equation}
    d \mbh_t  = f_\tf(\mbh_t, t) dt + G(\mbh_t,t) dW_t,\quad
    \theta_f \sim p_\phi(\tf), \label{eq:bbsde}
\end{equation}
which we refer to as a {\it Bayesian Neural Stochastic Differential Equation (BNSDE)}. The epistemic uncertainty introduced on the network weights allows the model to quantify the model uncertainty, i.e. the knowledge of which synaptic map fits best to data, in addition to the aleatoric uncertainty that the Wiener Process models. For technical reasons to be clarified below, we assume $f_\tf(\cdot,\cdot)$ and $G(\cdot,\cdot)$ to be $L$-Lipschitz-continuous, and $G(\cdot,\cdot)$ not to have any learnable parameters.

A coarse description of the environment dynamics is sometimes available as an incomplete set of differential equations in real-world applications. For instance, the dynamics of a three-dimensional volume might be modelled as a flow through a single point, such as the center of mass. Alternatively, a model on a subset of the system components might be provided. We assume this prior knowledge to be available as an ODE 
\begin{equation}
d{\bf h}_t = r_{\xi}({\bf h}_t,t)dt, \label{eq:prior-ode}
\end{equation}
where $r_\xi(\cdot,\cdot): \mathds{R}^P\times\mathds{R}_+\to \mathds{R}^P$ is an arbitrary non-linear function parameterized by a fixed set of parameters $\xi$. We can incorporate these known dynamics into the BNSDE by adding them to the drift as follows:
\begin{equation}
d{\bf h}_t =\big(f_{ \tf }\big({\bf h}_t,t)+\boldsymbol{\gamma} \circ r_{ \xi}({\bf h}_t,t)\big) dt + G({\bf h}_t,t) d{W}_t,
\label{eq:hybrid_sde}
\end{equation}
which can be viewed as a hybrid SDE with the free parameter vector $\boldsymbol{\gamma} \in [0,1]^P$ governing the relative importance of prior knowledge on the learning problem and $\circ$ referring to element-wise multiplication. Although we specified~\eqref{eq:prior-ode} within the same dimensional state space as~\eqref{eq:hybrid_sde}, $\boldsymbol{\gamma}$ allows us to provide only partial information. When prior knowledge is available only for a subset of the state space dimensions, the remaining dimensions $d$ can be filled in by simply setting $\gamma_d=0$.

We define a prior stochastic process representing solely the prior knowledge of the dynamics as
\begin{equation}
    d{\bf h}_t  = \big(\boldsymbol{\gamma} \circ r_{ \xi}({\bf h}_t,t)\big) dt + G({\bf h}_t,t) d{W}_t. \label{eq:prior_sde}
\end{equation}
This prior SDE will be used as a reference distribution for complexity penalization as part of the final PAC training objective of our hybrid SDE. Note that we have used the same diffusion term as in \eqref{eq:hybrid_sde} for specifying the prior SDE, which makes the complexity term within the PAC-formulation tractable, as we will show later. Also note that  $\mbgamma$ is a free parameter of the prior.

\subsection{Learning via Empirical Bayes}
 Solving the SDE in \eqref{eq:hybrid_sde} even for fixed parameters $\tf$ over an interval $[0,T]$ is analytically intractable for basically all practically interesting use cases. While our method is applicable to any discretization scheme, we demonstrate its use with the straightforward EM  for simplicity, which gives us the discrete-time version of the hybrid BNSDE below
\begin{align*}
    &\tf \sim p_\phi(\tf),\quad \mbh_0 \sim p(\mbh_0),\\
    &\mbh_{k+1}|\mbh_k,\tf \sim \Norm\left(\mbh_{k+1}\big|\mbh_k + d(\mbh_k ,t_k)\Delta t_k,\Sigma_k\right),\\
    &d(\mbh_k,t_k) = f_\tf(\mbh_k, t_k)+\boldsymbol{\gamma} \circ r_{ \xi}(\mbh_k,t_k),
\end{align*}
with $\Sigma_k:= \mbJ_k \Delta t_k, \mbJ_k:= G(\mbh_k,t_k)G(\mbh_k,t_k)^\top$, $\Delta t_k := t_{k+1} - t_k$, and $p(\mbh_0)$ defined on the initial state.

Analogously to latent state space models, we assume that the observations of the dynamics described in~\eqref{eq:bbsde}, \eqref{eq:hybrid_sde}, and \eqref{eq:prior_sde} are linked via a likelihood $p(\mby_k|\mbh_k)$. Specifically, we observe these dynamics as a time series $\mbY = \{\mby_1,\ldots, \mby_K\}$ consisting of $K$ observations $\mby_k \in \mdR^D$, collected at irregular time points $\mbt = \{t_1,\ldots,t_K\}$. 

Given an observed set of $N$ such time series trajectories ${\mcD = \{\mbY_1,\ldots,\mbY_N\}}$, the classical approach \citep{mackay2003information, gelman2013bayesian} would now require as a first step the inference of the posterior over both the global variables $\tf$ as well as the local variables $\mbH_n = \{\mbh_1^n,...,\mbh_K^n\}$, i.e.\ of $p(\tf,\mbH_1,\ldots, \mbH_N|\mcD)$, and as a second step a marginalization over this posterior to get the posterior predictive. As an analytical solution is intractable, approximate solutions such as Markov Chain Monte Carlo~(MCMC) methods or Variational Inference (VI) are required. Application of either of these approaches to BNSDEs is prohibitive, the former computationally, the latter in terms of expressiveness since existing work makes strong independence and structural assumptions on the approximate posterior. 

We propose in the following to apply model selection as an alternative path to BNSDE inference. Instead of performing the posterior inference on the latent variables, we marginalize them out and learn those hyperparameters $\phi$ from data that provide the highest log marginal likelihood~\citep{williams2006gaussian}. That is our BNSDE learns the optimal $\phi^*$ via 
\begin{align}
\argmax_{\phi} &\int  p(\mathcal{D}|\mbH) p(\mbH|\tf) p_\phi(\tf)d(\mbH,\tf) .\label{eq:marglik-ideal}
\end{align}
An advantage of this construction is that the marginal likelihood has the identical functional form to the predictive distribution, which is the quantity of interest in a typical prediction task. Marginal likelihood learning has also been applied before in the context of neural networks~\citep{sensoy2018evidential,malinin2018prior, garnelo18conditional}. Fitting the hyperparameters of an SDE to data via marginal likelihood maximization can also be viewed as an instance of the simulated likelihood method~\citep{sarkka2019applied}.

Marginalizing over $\tf$ in \eqref{eq:marglik-ideal} is intractable for most practical use cases. However, it can be approximated by Monte Carlo integration without constructing chains on the global parameters. Sampling directly from the prior, we get for a single observation~$n$ and $s=1,\ldots S$
\begin{equation}
\begin{split}
       &\theta_f^s \sim p_\phi(\tf),\quad \mbH^s \sim p(\mbH|\theta_f^s),\\
       &\phi^* := \argmax_{\phi}\log \Big(\frac1S \sum_{s=1}^S p(\mcD|\mbH^s)\Big). %
\end{split}
\label{eq:marglik-mc}
\end{equation}
In order to maximize this objective, we require an efficient computation of gradients w.r.t.\ the hyperparameters $\phi$. Access to $\phi$ is only given via samples from the distribution it is parameterizing. In our experiments, we assume this distribution $p_\phi(\tf)$ to be normal, allowing us to make use of a standard reparameterization. We  separate the sampling process into a parameter-free source of randomness and a parametric transformation, i.e.\ we have $\veps \sim p(\veps)$, $\tf = g_\phi(\veps)$, for a suitable $g_\phi(\cdot)$. In order to further reduce the variance noise introduced to the gradients due to this sampling step, we also use the \emph{local reparameterization trick}~\citep{kingma2015variational} in the drift, i.e.\ we sample the layer outputs during the forward propagation instead of individual layer weights. 

The objective~\eqref{eq:marglik-mc} is agnostic to the specific SDE employed. Therefore, we refer to the  discretized black-box SDE in~\eqref{eq:bbsde} governing $p(\mbH|\tf)$ and trained w.r.t.\ $\phi$ via this objective as \emph{E-Bayes} throughout the experiments. Analogously, we refer to training a hybrid SDE as in~\eqref{eq:hybrid_sde} with the same method as \emph{E-Bayes-Hybrid}.

\subsection{A Trainable PAC Bound}
A major downside of the objective in~\eqref{eq:marglik-mc}, when applied to BNSDEs, is that it optimizes a large set of hyperparameters, i.e.\ means and variances of drift network weights, without a proper regularization aside from the implicit regularization inherent in the chosen architecture and the marginalization itself. While the hybrid approach already allows us to incorporate prior expert knowledge, it remains a guiding signal without an explicit model capacity regularizer. Next, we address this problem by developing a training objective derived from a PAC-Bayesian bound objective that combines the benefits from the results we arrived at so far with a proper regularization scheme. 

The proposed approach is still agnostic to the chosen discretization scheme. Consequently, we refer for any time horizon $T > 0$ to  all local latent variables by $\mbh_{0\to T}$. To distinguish the density given by the hybrid SDE in~\eqref{eq:hybrid_sde} from the prior SDE in~\eqref{eq:prior_sde}, we further refer to the two densities induced by them respectively as $p_\text{hyb}(\mbh_{0\to T}|\tf)$ and $p_\text{pri}(\mbh_{0\to T})$. We define two distributions $Q$ and $P$ over $(\mbh_{0\to T}, \tf)$. For the former, we have the joint distribution of the hybrid process
\begin{equation}
    Q_{0\to T}(\mbh_{0\to T}, \tf) =p_\text{hyb}(\mbh_{0\to T}|\tf)p_\phi(\tf), 
\label{eq:hybridjoint}
\end{equation}
while the latter stands for the joint of the prior process%
\begin{equation}
    P_{0\to T}(\mbh_{0\to T}, \tf) =p_\text{pri}(\mbh_{0\to T}) p_\text{pri}(\tf). 
\end{equation}
Although the prior process is independent of the drift parameters $\tf$, we specify a fixed prior distribution $p_\text{pri}(\tf)$, which we choose to be a standard normal within our experiments. To be compliant with the notational practice in the PAC-Bayesian literature, we denote the prior distribution as $P$ and the posterior distribution that is fit to data as $Q$.\footnote{In the PAC-Bayesian framework, $P$ and $Q$ do not have to be linked to each other via application of the Bayes rule on an explicitly defined likelihood.}

As both $Q$ and $P$ share the same diffusion term, the Kullback-Leibler (KL) divergence between these processes can be calculated by extending the proof of~\citet{archambeau08variational}. The following Lemma holds for any choice of diffusion $G(\cdot,\cdot)$. You can find the proofs for it and the following Theorems in the appendix.

\paragraph{Lemma 1.} {\it For the process distributions} $Q_{0 \rightarrow T}$ {\it and} $P_{0 \rightarrow T}${\it, it holds that}
  \begin{align*}
      &D_{KL} \big(Q_{0 \rightarrow T} || P_{0 \rightarrow T}\big) =\\
      &~~~~~~~~\dfrac{1}{2} \int_{0}^T \mdE_{Q_{0 \rightarrow T} } \Big [ f_{ \tf }({\bf h}_t,t)^\top \mathbf{J}_t^{-1} f_{ \tf }({\bf h}_t,t) \Big ] dt \nonumber\\
          &~~~~~~~~~~~~~~~~~+ D_{KL}\big( p_\phi(\tf) || p_\text{pri}(\tf)\big ), \nonumber
  \end{align*}
 {\it for some $T>0$}, {\it where} $\mathbf{J}_t=  G({\bf h}_t,t) G({\bf h}_t,t)^\top$.

This Lemma\footnote{We assume $p_\text{pri}$ and $p_\phi$ to be Gaussians. However, to express the KL divergence between prior
and posterior processes analytically, it is sufficient for them to share the same diffusion. Following the established
definition of a stochastic process on which the It\^o calculus has been built, we assume Gaussian diffusion noise across
time increments. In our case, these Gaussian increments are warped by non-linear drift functions (neural nets) in the
subsequent time steps. Hence, they are capable of expressing arbitrarily complicated marginal process densities. In
effect, one can attain a Gaussian distributed marginal process only from a linear time-invariant SDE. The integral
 could be analogously defined also for other consistent increment choices (e.g., Levy-Flights instead of
Brownian motions).} provides one of the main ingredients for deriving a PAC-Bayesian bound on the generalization performance of a learned distribution $Q_{0 \rightarrow T}$. To derive such a bound, we additionally specify the risk via a loss function measuring the model mismatch. We assume the  likelihood  function $p({\bf y}_t|{\bf h}_t)$ to be uniformly bounded everywhere.\footnote{In our experiments, we ensure this condition by choosing the likelihood to be a normal density with bounded variance, i.e. bounded mass on the mode.} We then define the true risk of a draw from $Q_{0 \rightarrow T}$  on an i.i.d.\ sampled trajectory ${{\bf Y} = \{ {\bf y}_{1}, \dots, {\bf y}_{K} \}}$ at discrete and potentially irregular time points $t_1, \dots, t_K$ drawn from an unknown ground-truth stochastic process $\mathfrak{G}(t)$ as the expected model misfit on the sample. Specifically, we define the risk over hypotheses $H=({\bf h}_{0 \rightarrow T},\tf)$  as follows:
\begin{equation}
R(H) = 1-\mdE_{{\bf Y} \sim \mathfrak{G}(t)} \Big [ \prod_{k=1}^K  {p( {\bf y}_{k}|{\bf h}_{k})}/\overline{B}  \Big ],  \label{eq:true_risk}
\end{equation}
for time horizon $T > 0$ and the corresponding empirical risk on a data set $\mathcal{D}=\{ {\bf Y}_1, \dots, {\bf Y}_N \}$ as
\begin{equation}
    R_\mcD(H) = 1-\frac{1}{N} \sum_{n=1}^N \Big [ \prod_{k=1}^K 
                {p( {\bf y}_{k}^{n}|{\bf h}_{k}^{n})}/{\overline{B}}  \Big ]. 
\end{equation}
Here, $\overline{B} := \max_{{\bf y}_{k}, {{\bf h}_{k}} } p({\bf y}_{k}|{{\bf h}_{k}}) $ is a uniform bound to guarantee a $[0,1]$-valued loss.

Next, we develop a PAC-Bayesian generalization bound building on these risk definitions. Furthermore, we upper bound it with a trainable objective.%

\paragraph{Theorem 1.}
{\it The expected true risk is bounded above with probability $\mdP\geq 1-\delta$, for $\delta \in (0,1]$ by:}
 \begin{align}
  &\Ep{H\sim Q_{0 \rightarrow T}}{R(H)} \nonumber\\
  &~\leq \Ep{H\sim Q_{0 \rightarrow T}}{R_{\mathcal{D}}(H)} + \mathcal{C}_\delta(Q_{0 \rightarrow T},P_{0 \rightarrow T}) \label{eq:pac_bound}\\
       &~\le - \frac{1}{N} \sum_{n=1}^N \log\left(\frac{1}{S}\sum_{s=1}^S \prod_{k=1}^K p( {\bf y}_{k}^{n} |{\bf h}_{k}^{n,s})   \right) \label{eq:empirical_bayes_training}\\
        &~~~~+  \underbrace{\mathcal{C}_{\delta/2}(Q_{0 \rightarrow T},P_{0 \rightarrow T}) +\sqrt{\frac{\log(2N/\delta)}{2S}} + K\log \overline{B}}_{=:C}, \nonumber\\
         &~\le - \frac{1}{SN} \sum_{n=1}^N \sum_{s=1}^S \sum_{k=1}^K    \log \Big(p( {\bf y}_{k}^{n} |{\bf h}_{k}^{n,s})   \Big) \label{eq:training_loss} +C
 \end{align}
{\it with  $S$ the sample count taken independently for each observed sequence, and the complexity functional:}
\begin{align*}
		\mathcal{C}_\delta(&H_{0 \rightarrow T},P_{0 \rightarrow T}) =\\
		&\sqrt{\frac{{D_{KL}\big(Q_{0 \rightarrow T} || P_{0 \rightarrow T}\big) + \log({2\sqrt{N}}) - \log( {\delta/2})}}{2N}}
\end{align*}
{\it where $D_{KL}\big(Q_{0 \rightarrow T} || P_{0 \rightarrow T}\big)$ as in Lemma 1.}

As the complexity term in~\eqref{eq:training_loss} vanishes for large sample sizes $(N, S)$, the first term converges to the expected
log-likelihood for a given time resolution $K$. Although the bound loosens as $K$ increases, note that the gradient of the
bound w.r.t.\ model parameters remains unaffected.
Theorem 1 can be used to learn a posterior distribution $Q_{0 \rightarrow T}$ from data by adjusting $\phi$. Additionally, we can also learn the importance of the prior by fitting the $\gamma$ parameter to data. While directly learning $\gamma$ by optimizing the PAC-bound violates the generalization guarantee, we can define a collection of prior distributions $P_{0 \rightarrow T}$ for a set $\Gamma$ of discretized values of $\gamma$ and employ the same union bound as \cite{reeb18pacgp}. The resulting PAC-bound differs by a constant accounting for the number of distinct $\gamma$ values within the collection. Therefore, we can use the same gradient based optimization to learn $\gamma$ and quantize the value to the closest point within $\Gamma$ to evaluate the PAC bound.

\subsection{The Training Algorithm}
The first term in~\eqref{eq:pac_bound} does not correspond to the Empirical Bayes objective as it averages over likelihoods, and not log-likelihoods \citep{germain16pac}. However, the first term in~\eqref{eq:empirical_bayes_training} provides a sampling based approximation to the empirical Bayes objective. By defining the risk in such a way and employing the PAC-Bayesian framework, we obtain a regularized version of empirical Bayes.
Although placing the $\log(\cdot)$ function into its summands loosens the bound on the true risk, it improves numerical robustness and optimizing \eqref{eq:training_loss} still tightens the original PAC-Bayesian bound, i.e. \eqref{eq:pac_bound}, as stated in the following corollary. 

\paragraph{Corollary 1.}
{\it For Lipschitz-continuous risk and likelihood, a gradient step that reduces~\eqref{eq:training_loss} also tightens the PAC bound in~\eqref{eq:pac_bound}. }

Minimizing~\eqref{eq:training_loss} hence  closes the loop as the Empirical Bayes objective derived in~\eqref{eq:marglik-mc} reappears in \eqref{eq:empirical_bayes_training} but is now combined in a principled way with the regularization term $\mcC_\delta$. We can ignore the terms that do not depend on $\phi$ and adopt the remaining expression bound as our final objective and learn $\phi^*$ via 
\begin{equation}
\begin{split}
    &\phi^* :=\argmax_{\phi}~ \frac{1}{SN} \sum_{n=1}^N \sum_{s=1}^S \sum_{k=1}^K  \log \Big(p( {\bf y}_{k}^{n} |{\bf h}_{k}^{n,s})  \Big) \\ 
         &~~+  \sqrt{\Big(D_{KL}\big(Q_{0 \rightarrow T} || P_{0 \rightarrow T}\big ) + \log(4\sqrt{N}/\delta)\Big)/2N}.
         \end{split}
\end{equation}
In this training procedure, we only train w.r.t.\  $\phi$ which determine the drift term. To also learn the diffusion, one could represent $G$ also by a BNN. However, the corresponding training procedure would invalidate the PAC statement. Nevertheless, the diffusion term could be learnt on a held-out data set and then incorporated as fixed to the bound~\eqref{eq:training_loss}. As Theorem~1 applies to any diffusion term, we keep the genericness of its statement. However, in the experiments, we stick to a constant diffusion term for practical reasons.

Although we require i.i.d.\ observations of time series in the theory, we can in practice use mini-batches of trajectories provided that the batches are sufficiently far apart so that they become essentially independent. The objective~\eqref{eq:empirical_bayes_training} differs from the Empirical Bayes one in~\eqref{eq:marglik-mc} only by the complexity term. The only complicated calculation step in this term is the integral through the process, which can be made more implementation friendly using Fubini's theorem:
\begin{align*}
    &\int_{0}^T \mdE_{Q_{0 \rightarrow T} } \Big [ f_{ \tf }({\bf h}_t,t)^\top \mathbf{J}_t^{-1} f_{ \tf }({\bf h}_t,t) \Big ] dt\\
    &~~~~~= \mdE_{Q_{0 \rightarrow T} } \Big [ \int_{0}^T   f_{ \tf }({\bf h}_t,t)^\top \mathbf{J}_t^{-1} f_{ \tf }({\bf h}_t,t) dt \Big ].
\end{align*}
A pseudo-code description of the  procedure is given in Algorithm~\ref{alg:bnsde}. Our sampling-based method naturally couples with the EM approximation and inherits its convergence properties. We show strong convergence to the true solution with shrinking step size by extending the plain EM proof \citep{kloeden2011numerical}.

\begin{algorithm}[t!]
    \DontPrintSemicolon
    \SetKwInput{Input}{Input}
    \SetKwInput{Output}{Output}

    \Input{set of $N$ trajectories $\mcD$, prior drift $r_\xi(\cdot,\cdot)$, time points $\mbt$, drift $f_\tf(\cdot,\cdot)$, diffusion $G(\cdot,\cdot)$, weight distribution $p_\phi(\tf)$, number of samples $S$, prior parameter~$\gamma$}
    \Output{training objective loss}
    \BlankLine
    \tcp{init.\ marginal log-likelihood (mll) and kl}
     $\text{mll} \leftarrow 0; \text{kl} \leftarrow 0$\;
     \For(\tcp*[h]{for each trajectory}){$n \in \{1,\ldots,N\}$ }{
         \For(\tcp*[h]{and each sample}){$s \in \{1,\ldots, S\}$} {
         \tcp{sample initial state and weights}
             $\mbh_0^{n,s}\sim p(\mbh_0)$;
             $\theta_f^{n,s}\sim p_\phi(\tf)$\;
             \tcp{for each of the $K$ steps}
             \For{$k \in \{1,\ldots, K\}$}{
                \tcp{get drift,prior,diffusion output}
                 $f_k^{n,s}\leftarrow f_{\theta_f^{n,s}}(\mbh_{k-1}^{n,s}, t_{k-1})$\;
                 $r_k^{n,s}\leftarrow r_{\xi}(\mbh_{k-1}^{n,s}, t_{k-1})$\;
                 $G_k^{n,s}\leftarrow G(\mbh_{k-1}^{n,s}, t_{k-1})$\;
                 \tcp{sample stochasticity}
                 $\Delta t_k \leftarrow t_k - t_{k-1}$\;
                 $W_k^{n,s} \sim \Norm(0, \Delta t_k\mathds{1})$\;
                 \tcp{update state}
                 $\mbh_k^{n,s}\leftarrow \mbh_{k-1}^{n,s} + (f_k^{n,s} + \mbgamma r_k^{n,s})\Delta t_k + G_k^{n,s}W_k^{n,s}$\;
                 \tcp{and update mll and kl}
                 $\text{mll} \leftarrow \text{mll} + \tfrac{1}{SN}\log p(\mby_k^n|\mbh_k^{n,s})$\;
                 $\text{kl}\leftarrow \text{kl} + \tfrac{1}{2S}f_k^{n,s}{}^\top (G_k^{n,s}G_k^{n,s}{}^\top)^{-1}f_k^{n,s}\Delta t_k$\;
             }
         }
     }
     \tcp{add penalty for modified drift distribution}
         $\text{kl}\leftarrow \text{kl} + D_{KL}\Big(p_\phi(\tf)||p_\text{pri}(\tf)\Big)$\;
         \tcp{and assign final loss}
         $\text{loss} \leftarrow -\text{mll} + \sqrt{\big(\text{kl} + \log(4\sqrt{N}/\delta)\big)/(2N) }$\;
    \tcp{to be returned and optimized}
    \Return{$\mathrm{loss}$}
    \BlankLine
    \caption{E-PAC-Bayes-Hybrid Loss}\label{alg:bnsde}
\end{algorithm}

\paragraph{Theorem 2 (strong convergence).} {\it Let ${\bf h}_t^{\theta}$ be an It\^{o} process as in~\eqref{eq:bbsde} with drift and diffusion parameters $\theta$ and $\widetilde{\bf h}_t^{\theta}$ its Euler-Maruyama approximation for some regular step size $\Delta t>0$. For some coefficient $R>0$ and any $T>0$, the below inequality holds as $S\to\infty$ }
\begin{equation*}
    \mathds{E} \Bigg [ \sup_{0 \leq t \leq T} \Big | \mathds{E}_{\theta}[{\bf h}_t^{\theta}] - \dfrac{1}{S} \sum_{s=1}^S \widetilde{\bf h}_t^{\theta^{(s)}} \Big | \Bigg ] \leq R \Delta t^{1/2},
\end{equation*}
{\it where $\theta^{(s)}$ are i.i.d.\ draws from a prior $p_{\phi}(\theta)$. }

\section{RELATED WORK}\label{sec:related-work}
\rdone{We will discuss all papers pointed out by the R's in two new paragraphs of Sec 5:
i) The relation between variational inference and PAC Bayes: We will cite Knoblauch et al. 2019 and relate it to the baseline Hegde et al. 2019.
ii) Black-box identification and ABC of nonlinear dynamical systems: While (Brunton, Durstewitz, Park) differ in the model class used for fitting the r.h.s. of the differential equation, they can be combined with our approach as they admit an explicitly defined transitional noise. This would result in an adapted loss function (modified likelihood and prior over parameter values). Kennedy \& O'Hagan is limited to stationary data and its extension to time series is not straightforward.:
added black blox section (relating it to hedge, mentioned knoblauch in empirical based}

\paragraph{Empirical Bayes as PAC Learning.} \citet{germain2016pac} propose a learnable PAC-Bayesian bound that provides generalization guarantees as a function of a marginal log-likelihood. Our method differs from this work in two main lines. First, \citet{germain2016pac} define risk as $-\log p({\bf Y}|{\bf H}) \in (-\infty, +\infty)$ and compensate for the unboundedness by either truncating the support of the likelihood function or introducing assumptions on the data distribution, such as sub-Gaussian or sub-Gamma. Our risk defined in~\eqref{eq:true_risk} assumes uniform boundedness, yet can be incorporated into a PAC-Bayesian bound without further restrictions. Second, \citet{germain2016pac}'s bound is an unparameterized rescaling of the marginal log-likelihood. Hence, it is not linked to a capacity penalizer, which can be used at {\it training time} for regularization. Applying this method to hybrid sequence modelling boils down to performing plain Empirical Bayes, i.e.\ \emph{E-Bayes} in our experiments. 

\paragraph{Differential GPs.} \citet{hegde19differential} model the  dynamics of the activation maps of a {\it feed-forward} learner by the predictive distribution of a GP. This method allocates the mean of a GP as the drift and covariance as the diffusion. It infers the resultant model using variational inference.  While direct application of this method to time series modeling is not straightforward, we represent it in our experiments by sticking to our generic non-linear BNSDE design in~\eqref{eq:bbsde}, and inferring it by maximizing the ELBO: $\mathcal{L}(\phi) = \mdE_{{\bf H}, \theta}\big[ \log~p({\bf Y}|\mbH)\big]-D_{KL}\big( p_{\phi}(\theta)||p(\theta)\big),$
applying the local reparameterization trick on $\theta$.
Although variational inference can be seen from a PAC-perspective by choosing the log-likelihood as the loss \citep{knoblauch2019generalized}, the ELBO does not account for the deviation of variational posterior over latent dynamics from the prior latent dynamics. We refer to this baseline in the experiments as \emph{D-BNN (VI)}. The approximate posterior design here closely follows the PR-SSM approach \citep{doerr18probabilistic}, which represents state of the art in state-space modelling.

\paragraph{Differential BNNs with SGLD.} The learning algorithm of \citet{look2019diff} shares our BNSDE modeling assumptions, however, it uses Stochastic Gradient Langevin Dynamics (SGLD) to infer $\theta$. The algorithm is equivalent to performing MAP estimation of the model parameters in~\eqref{eq:bbsde} while distorting the gradient updates with decaying normal noise that also determines the learning rate. 

\paragraph{Black-box identification of dynamic systems.} There are various approaches to identify a dynamical system that differ in the model class used for fitting the right-hand side of the differential equation and may also allow for transitional noise \citep[e.g.][]{brunton2016discovering,durstewitz2016state}. These approaches could be incorporated into ours, using their transition likelihood and prior over parameters. Our black-box neural SDE can be seen as one instance of such a black-box identification of dynamical systems ({\em{E-Bayes}}). As we are mainly interested in incorporating prior knowledge into such black-box models, we chose one such competitor \citep{hegde19differential},
with reported results on the CMU Motion capture data set (Tab.~\ref{tab:cmu}).

\section{EXPERIMENTS}\label{sec:experiments}

We evaluate the following four variants of our method:
\begin{enumerate}
    \item[(i)] \emph{E-Bayes}.  Empirical Bayes without prior knowledge, i.e.\ training~\eqref{eq:marglik-mc} with  $p({\bf h}_{0 \rightarrow T})$ given by \eqref{eq:bbsde}. 
    \item[(ii)] \emph{E-PAC-Bayes.} Empirical PAC Bayes on the BNDSE using the objective in~\eqref{eq:training_loss} with an uninformative prior drift, i.e.\ $r_{\xi}({\bf h}_t,t) = 0$.
    \item[(iii)] \emph{E-Bayes-Hybrid.} Same training objective as (i), however with the hybrid model as proposed in \eqref{eq:hybrid_sde}.
    \item[(iv)] \emph{E-PAC-Bayes-Hybrid}. The hybrid model~\eqref{eq:hybrid_sde} with the same loss as \emph{E-PAC-Bayes}, which is the combination we propose.
\end{enumerate}
We extend the Empirical Bayes objective in~\eqref{eq:marglik-mc} by PAC-Bayes to tune many hyperparameters without overfitting and incorporate prior domain knowledge in a principled way. We evaluate the first motivation as \emph{E-PAC-Bayes}, i.e.\ objective~\eqref{eq:training_loss} but without a prior SDE, and the complete model including a prior SDE as \emph{E-PAC-Bayes-Hybrid}. See the appendix for a detailed discussion of each of these methods' computational cost and further experiments. 

\begin{table}
    \centering
    \caption{Ablation study on the Lorenz attractor to evaluate the contributions of the prior knowledge on the predictive performance measured in Mean Squared Error (MSE) with standard error over fifty repetitions. The hybrid models ({(iii), (iv)}) consistently improve on the black box models ({(i),(ii)}). The last row (v) shows the performance for the case the model has full access to the true dynamics with noisy parameters in~\eqref{eq:prior-ode}.}
    \begin{adjustbox}{max width=0.95\linewidth}
    \begin{tabular}{ccc}
    \toprule
    \textbf{Prior Knowledge} & \textbf{Model} &  \textbf{Test MSE} \\
    \midrule
         \multirow{2}{*}{None} & {(i)} & $29.20 \pm 0.19$ \\
          & {(ii)} & $29.05 \pm 0.23$\\ \midrule
        \multirow{2}{*}{$\boldsymbol{\gamma}=[1, 0, 0],~~~\zeta \sim \mathcal{N}(10,1)$}& {(iii)}& $27.58 \pm 0.17$\\
        & {(iv)} & $27.42 \pm 0.16$\\\midrule
        \multirow{2}{*}{$\boldsymbol{\gamma}=[0, 1, 0],~~~\kappa \sim \mathcal{N}(2.67,1)$}& {(iii)} & $15.87 \pm 0.46 $ \\
        & {(iv)} & $15.06 \pm 0.35$\\\midrule
        \multirow{2}{*}{$\boldsymbol{\gamma}=[0, 0, 1],~~~\rho \sim \mathcal{N}(28,1)$}& {(iii)} & $27.82 \pm 0.26$ \\
        & {(iv)} & $28.37 \pm 0.21$\\\midrule
        $\boldsymbol{\gamma}=[1,1,1],$ & \multirow{2}{*}{(v)} & \multirow{2}{*}{$16.40 \pm 2.31$}\\
        $ (\zeta, \kappa, \rho)^\top \sim \Norm\big((10, 2.67, 28)^\top, \mathds{1}_3\big)$ & & \\
         \bottomrule
    \end{tabular}
    \end{adjustbox}
    \label{tab:lorenz}
\end{table}

\rdone{1d trajectory visualizations of the Lorenz attractor // prob for the appendix with 3d in the main paper}

\begin{table*}[t]
	\caption{Benchmarking of our method on the CMU Motion Capture Data Set. Mean Squared Error (MSE) and Negative Log-Likelihood (NLL) on $300$ future frames is averaged over ten repetitions ($\pm$ standard deviation).}
	\label{tab:cmu}
\centering
\begin{adjustbox}{max width=0.95\linewidth}
			\begin{tabular}{lcccccc}
				\toprule
				{\bf Method} &  {\bf Reference} & {\bf Bayesian} & {\bf Hybrid} & {\bf +KL} & {\bf Test MSE} & {\bf Test NLL}  \\
				\midrule
				$\text{DTSBN-S}$ & \citep{gan2015deep}           & No & No & No & $34.86 \pm 0.02$   & Not Applicable \\
 				$\text{npODE}$ & \citep{heinonen2018learning}              & No & No & No & $22.96$ & Not Applicable \\
				$\text{Neural-ODE}$ & \citep{chen2018neural}         & No & No & No & $22.49 \pm 0.88$ & Not Applicable \\
				$\text{ODE}^2\text{VAE}$ & \citep{yildiz19odevae} & Yes & Yes & No & $10.06 \pm 1.40$ & Not Reported\\ 
		  	    $\text{ODE}^2\text{VAE-KL}$ & \citep{yildiz19odevae} & Yes & Yes & Yes & $8.09 \pm 1.95$ & Not Reported \\
				D-BNN (SGLD) & \citep{look2019diff} & Yes & No & No &$13.89 \pm 2.56$ & $747.92 \pm 58.49$\\
				D-BNN (VI) & \citep{hegde19differential} & Yes & No & Yes & $9.05 \pm 2.05$ & $452.47 \pm 102.59$\\
				\midrule
				$\text{E-Bayes}$ &  Baseline           & Yes & No & No & $8.68 \pm 1.56$ & $433.76 \pm 77.78$\\
				\midrule
				$\text{E-PAC-Bayes}$ &  Ablation & Yes & No & Yes & $9.17 \pm 1.20$ & $489.82 \pm 67.06$ \\
				$\text{E-Bayes-Hybrid}$ &  Ablation & Yes & Yes & No & $9.25 \pm 1.99$ & $462.82 \pm 99.61$ \\
				$\text{E-PAC-Bayes-Hybrid}$  &  Proposed & Yes & Yes & Yes & ${\bf 7.84 \pm 1.41} $ & ${\bf 415.38 \pm 80.37}$ \\
				\bottomrule
			\end{tabular}
\end{adjustbox}
\end{table*}

\paragraph{Lorenz Attractor.}
This chaotic non-linear system has the the following inherently unsolvable dynamics
\begin{align*}
    dx_t &= \zeta (y_t-x_t)dt + dW_t, \\
    dy_t &= \big(x_t (\kappa-z_t)-y_t\big)dt + dW_t, \\
    dz_t &= (x_t y_t - \rho z_t)dt + dW_t,
\end{align*}
where $\zeta=10, \kappa=28, \rho=2.67$, and $W_t$ is a random variable following a Wiener process with unit diffusion. We generate 1920 observations from the above dynamics initiating the system at $x_0,y_0,z_0 = (1,1,28)$, use the first half for training and the rest for testing.  We split both the training and the test data into 20 sequences of length 24, which can be interpreted as i.i.d.\ samples of the system with different initial states. Table~\ref{tab:lorenz} presents the 24-step ahead forecasting error in MSE on the test set for our model variants. In each experiment repetition, E-Bayes-Hybrid and E-PAC-Bayes-Hybrid are provided one equation after distorting the corresponding parameter by normal distributed noise. The other equations are hidden by being hard assigned to zero. To set up the corresponding prior and model, we used a constant diffusion with $G=\mathds{1}$.
Despite the imprecision of the provided prior knowledge, the largest performance leap comes from the hybrid models. The complexity term on the PAC-Bayesian bound restricts the model capacity for black-box system identification, while it improves the hybrid setup.

\begin{figure}[ht!]
      \centering
        \includegraphics[width=0.8\linewidth]{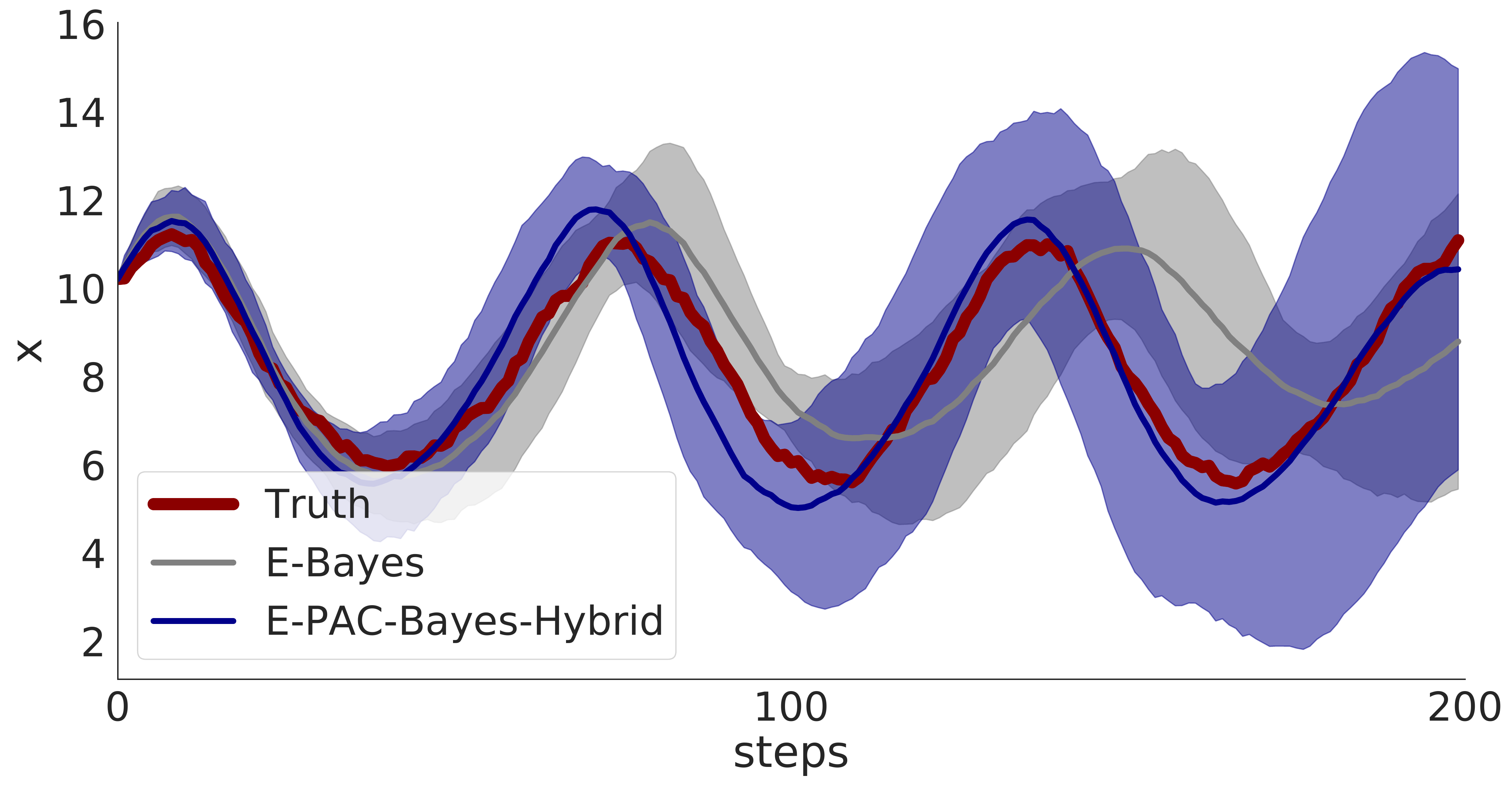}
    \caption{Predicted mean trajectory starting at $T=10$ on one dimension of the Lorenz data. The shaded areas give $\pm 2$ standard deviations over 21 trajectories.}
    \label{fig:lorenz1d}
\end{figure}

Figure~\ref{fig:lorenz1d} visualizes the predicted trajectories on the test sequence for prior knowledge on $dz_t$. Even with weak prior knowledge, the proposed model is stable longer than the baseline and shows a proper increase in the predictive variance over time.

\paragraph{CMU Walking Data Set.}
We benchmark against state of the art on this motion capture data set following the setup of \citet{yildiz19odevae}. We train an \emph{E-PAC-Bayes} model on the \texttt{MOCAP-1} data set consisting of 43 motion capture sequences measured from 43 different subjects. The drift net of the learned BNSDE is then treated as weak and broad prior knowledge of human walking dynamics. We use \texttt{MOCAP-2} with 23 walking sequences from Subject 35 to represent a high-fidelity subject-specific modelling task. As reported in Table 2 of \citet{yildiz19odevae}, the state of the art of subject-independent mocap dynamic modelling has twice as high prediction error as subject-specific dynamics (MSE of $15.99$ versus $8.09$). Analogously to the Lorenz attractor experiment, we fixed the PAC-variants' prior diffusion term to be constant. We report the test MSE and negative log-likelihoods in Table \ref{tab:cmu}. Our method delivers the best prediction accuracy and model fit when all its components are active.

\section{CONCLUSION}\label{sec:conclusion}
We have shown that our method incorporates vague prior knowledge into a flexible Bayesian black-box modelling approach for learning SDEs resulting in a robust learning scheme guided by generalization performance via a PAC-Bayesian bound. The method is easily adaptable to other solvers. For example, the training loss derived in~\eqref{eq:marglik-mc} can also be optimized using a closed-form normal assumed density scheme applied over a stochastic Runge-Kutta variant \citep{li2019stochastic}. 
Independent from the sampling scheme and model used, our tied gradient update procedure allows training on the loose, yet numerically stable, bound while providing an improvement w.r.t.\ the generalization guarantees on its tighter counterpart. Our stochastic approximation of the data log-likelihood currently relies on samples obtained from the prior, yet could be improved by incorporating a more sophisticated sampling scheme, e.g.\ using particle filtering~\citep{kantas2015particle}. Finally, the bound  in~\eqref{eq:training_loss} has the potential to be vacuous for certain drift nets, incorporating a Hoeffding assumption \citep{alquier2016properties} could further tighten it.

\bibliographystyle{plainnat} 
\bibliography{main}

\onecolumn
\begin{center}
    \textbf{\Large{Learning Partially Known Stochastic Dynamics \\ with Empirical PAC Bayes\\ ---\textsc{APPENDIX}---}}
\end{center}
\runningtitle{Learning Partially Known Stochastic Dynamics with Empirical PAC Bayes --- Appendix}
\section{CONTINUOUS TIME SDES}\label{app:continuous_time_sde}

Solving the SDE system in~(1)
for a time interval $[0,T]$ and fixed $\theta_f$ requires computing integrals of the form
\begin{equation*}
 \int_0^T d{\bf h}_t =  \int_0^T f_{ \theta_f }({\bf h}_t,t) dt +  \int_0^T G({\bf h}_t,t) dW_t.
 \end{equation*}
This operation is intractable for almost any practical choice of $f_{ \theta_f }(\cdot,\cdot)$ and $G(\cdot,\cdot)$ for two reasons. First, the integral around the drift term $f_{ \theta_f }(\cdot,\cdot)$ does not have an analytical solution, due both to potential nonlinearities of the drift and to the fact that ${\bf h}_t \sim p({\bf h}_t,t)$ is a stochastic variable following an implicitly defined distribution. Second, the diffusion term involves the It\^{o} integral \citep{oksendal92stochastic} about $W_t$ which multiplies the non-linear function $G(\cdot,\cdot)$.

For each of the SDEs in 
(6) and (7),
we could alternatively to the Euler-Maruyama integration theme use the  Fokker-Planck-Kolmogorov equation to derive a partial differential equation (PDE) system 
\begin{align*}
 \partial p_\text{hyb}({\bf h}_t,t|\theta_f)/\partial t &= 
 -\nabla \cdot \big[\big(f_{ \theta_f }({\bf h}_t,t)+\boldsymbol{\gamma} \circ r_{ \xi}({\bf h}_t,t)\big) p_\text{hyb}({\bf h}_t,t|\theta_f)\big]
 + \nabla \cdot \big(\mathbf{1} \nabla \cdot  G({\bf h}_t,t) p_\text{hyb}({\bf h}_t,t|\theta_f)\big),\\
 \partial p_\text{pri}({\bf h}_t,t)/\partial t &= -\nabla \cdot \big[\big(\boldsymbol{\gamma} \circ r_{ \xi}({\bf h}_t,t)\big)p_\text{pri}({\bf h}_t,t)\big] 
 + \nabla \cdot \big(\mathbf{1} \nabla \cdot   G({\bf h}_t,t)p_\text{pri}({\bf h}_t,t)\big),
\end{align*}
where $\nabla \cdot$ is the divergence operator and $\mathbf{1} =  (1,\dots,1)^\top$. 
Theoretically, these distributions can be obtained by solving the Fokker-Planck PDE. As this requires solving a PDE which is not analytically tractable, we instead resort to the discrete time Euler-Maruyama integration.
\section{PROOFS}\label{app:proofs}
\rtodo{fix margin breaks}

This section gives a more detailed derivation of the individual results stated in the main paper.

\paragraph{Lemma 1.\label{lem:sde_kl}} {\it For the process distributions}\footnote{See the main paper for their definitions.} $Q_{0 \rightarrow T}$ {\it and} $P_{0 \rightarrow T}$ {\it the following property holds}
 \begin{align*}
     D_{KL} \big(Q_{0 \rightarrow T} || P_{0 \rightarrow T}\big) &= \dfrac{1}{2} \int_{0}^T \mdE_{Q_{0 \rightarrow T} } \Big [ f_{ \theta_f }({\bf h}_t,t)^\top \mathbf{J}_t^{-1} f_{ \theta_f }({\bf h}_t,t) \Big ] dt +D_{KL}\big( p_\phi(\theta_f) || p_\text{pri}(\theta_f)\big )
 \end{align*}
 {\it for some $T>0$}, {\it where} $\mathbf{J}_t=  G({\bf h}_t,t) G({\bf h}_t,t)^\top$ .

\paragraph{Proof. } Assume Euler-Maruyama discretization for the process $Q_{0 \rightarrow T}$ on arbitrarily chosen $K$ time points within the interval $[0,T]$. Then we have $D_{KL}(Q || P)$ denoting the Kullback-Leibler divergence between processes $Q_{0 \rightarrow T}$ and $P_{0 \rightarrow T}$ up to discretization into $T$ time points as:
\begin{align*}
  D_{KL}(Q || P) &=
	\iint \log \dfrac{  \prod_{t=0}^{K-1} \Big (  \mathcal{N}\big({\bf h}_{t+1}|\left(f_{\theta_f}({\bf h}_t,t)+\boldsymbol{\gamma}\circ r_{\xi}({\bf h}_t,t)\right)\Delta t , \mathbf{J}_t \Delta t \big) \Big )  }{\prod_{t=0}^{K-1} \Big ( \mathcal{N}\big({\bf h}_{t+1}|\boldsymbol{\gamma}\circ r_{\xi}({\bf h}_t,t)\Delta t , \mathbf{J}_t \Delta t \big) \Big ) }\cdot  \frac{\cancel{p({\bf h}_0)} p_\phi(\theta_f) }{\cancel{p({\bf h}_0)} p_\text{pri}(\theta_f)} Q_{0 \rightarrow T} d{\bf H} d\theta_f \\
	&=\sum_{t=0}^{K-1} \iint  \log \mathcal{N}\big({\bf h}_{t+1}|\left(f_{\theta_f}({\bf h}_t,t)+\boldsymbol{\gamma}\circ r_{\xi}({\bf h}_t,t)\right)\Delta t , \mathbf{J}_t \Delta t \big) \\
	&\qquad\qquad\quad -\log\mathcal{N}\big({\bf h}_{t+1}|\boldsymbol{\gamma}\circ r_{\xi}({\bf h}_t,t)\Delta t , \mathbf{J}_t \Delta t \big) Q_{0 \rightarrow T}  d{\bf H} d\theta_f \\
	&\qquad+D_{KL}\big(p_\phi(\theta_f)||p_\text{pri}(\theta_f)\big).
\end{align*}
For simplicity, let us modify notation and adopt ${\bf f}_t:=f_{\theta_f}({\bf h}_t,t)+\boldsymbol{\gamma}\circ r_{\xi}(\mathbf{h}_t,t)$, ${\bf g}_t:=\boldsymbol{\gamma}\circ r_{\xi}(\mathbf{h}_t,t)$, and ${\Delta {\bf h}_{t+1} := {\bf h}_{t+1}-{\bf h}_t}$. Now writing down the $\log(\cdot)$ terms explicitly, we get
\begin{align*}
  D_{KL}(Q || P) &= \frac{1}{2} \sum_{t=0}^{K-1} \iiint  \Big [-(\Delta {\bf h}_{t+1}-{\bf f}_t\Delta t)^\top (\mathbf{J}_t \Delta t)^{-1} (\Delta {\bf h}_{t+1}-{\bf f}_t\Delta t)\\
&\qquad\qquad\qquad +(\Delta {\bf h}_{t+1}-{\bf g}_t\Delta t)^\top (\mathbf{J}_t \Delta t)^{-1} (\Delta {\bf h}_{t+1}-{\bf g}_t\Delta t) \Big ]  \\
&\qquad\qquad\qquad  \cdot p_\text{hyb}({\bf h}_{0 \rightarrow T}|\theta_f) p_{\phi}(\theta_f)    d{\bf H} d\theta_f\\
&\qquad +D_{KL}\big(p_\phi(\theta_f)||p_\text{pri}(\theta_f)\big).
\end{align*}
Expanding the products, removing the terms that cancel out, and rearranging the rest, we get
\begin{align*}  
  D_{KL}(Q || P) &=\frac{1}{2}\sum_{t=0}^{K-1}  \iint \Big [ -{\bf f}_t^\top \mathbf{J}_t^{-1} {\bf f}_t \Delta t + 2 \Delta {\bf h}_{t+1} \mathbf{J}_t^{-1} {\bf f}_t + {\bf g}_t^\top \mathbf{J}_t^{-1} {\bf g}_t \Delta t - 2 \Delta {\bf h}_{t+1} \mathbf{J}_t^{-1} {\bf g}_t \Big ] \\
	&\qquad\qquad\qquad\cdot p_\text{hyb}({\bf h}_{0 \rightarrow T}|\theta_f,) p_{\phi}(\theta_f)    d{\bf H} d\theta_f \\
	&\qquad +D_{KL}\big(p_\phi(\theta_f)||p(\theta_f)\big).
\end{align*}
Note that from the definition of the process it follows that
\begin{equation*}
\int \Delta {\bf h}_{t+1}~~ p_\text{hyb}({\bf h}_{0 \rightarrow T}|\theta_f) d \Delta {\bf h}_{t+1} = {\bf f}_t \Delta t.
\end{equation*}
Plugging this fact into the KL term, we have
\begin{equation*}  
  D_{KL}(Q || P) =\dfrac{1}{2}\sum_{t=0}^{K-1} \int  \Big [ {\bf f}_t^\top \mathbf{J}_t^{-1} {\bf f}_t \Delta t + {\bf g}_t^\top \mathbf{J}_t^{-1} {\bf g}_t \Delta t - 2 {\bf f}_t \mathbf{J}_t^{-1} {\bf g}_t \Delta t \Big ] p_{\phi_f}(\theta_f) )  d\theta_f +D_{KL}\big(p_\phi(\theta_f)||p_\text{pri}(\theta_f)\big).
\end{equation*}
For any pair of vectors ${\bf a}, {\bf b} \in \mdR^P$ and symmetric matrix ${\bf C} \in \mdR^{P \times P}$, the following identity holds:
\begin{equation*}
{\bf a}^\top {\bf C} {\bf a} - {\bf b}^\top {\bf C} {\bf b} = ({\bf a}-{\bf b})^\top {\bf C} ({\bf a}-{\bf b}) + 2 {\bf a}^\top {\bf C} {\bf b}.
\end{equation*}
Applying this identity to the above, we attain
\begin{equation*}  
  D_{KL}(Q || P) =\dfrac{1}{2}\sum_{t=0}^{K-1} \int  \Big [ ({\bf f}_t-{\bf g}_t)^\top \mathbf{J}_t^{-1} ({\bf f}_t-{\bf g}_t) \Delta t \Big ] p_{\phi_f}(\theta_f)  d\theta_f +D_{KL}\big(q(\theta_f)||p(\theta_f)\big).
\end{equation*}
Plugging back the original terms and setting $K$ to the limit, we arrive at the desired outcome
\begin{align*}  
  &\lim_{K \rightarrow +\infty} \Bigg \{\frac{1}{2}\sum_{t=0}^{K-1}  \int  \Big [ (f_{\theta_f}({\bf h}_t,t))^\top \mathbf{J}_t^{-1} f_{\theta_f}({\bf h}_t,t) \Delta t \Big ] p_{\phi_f}(\theta_f)  d\theta_f  +D_{KL}\big(p_\phi(\theta_f)||p_\text{pri}(\theta_f)\big) \Bigg \}\\
	&\qquad= \frac{1}{2}\int \Big [ \int  f_{\theta_f}({\bf h}_t,t)^\top \mathbf{J}_t^{-1} f_{\theta_f}({\bf h}_t,t)  p_{\phi_f}(\theta_f)   d\theta_f  \Big ] dt+D_{KL}\big(p_\phi(\theta_f)||p_\text{pri}(\theta_f)\big)\\
         &\qquad=\frac{1}{2} \int_{0}^T \mathbb{E}_{Q_{0 \rightarrow T} } \Big [ f_{ \theta_f }({\bf h}_t,t)^\top \mathbf{J}_t^{-1} f_{ \theta_f }({\bf h}_t,t) \Big ] dt+ D_{KL}\big( p_\phi(\theta_f) || p_\text{pri}(\theta_f)\big ).	
\end{align*}\hfill\qed

\paragraph{Theorem 1.} {\it Let $p({\bf y}_t|{\bf h}_t)$ be uniformly bounded likelihood function with density $p({\bf y}_t|{\bf h}_t)$ everywhere and $Q_{0 \rightarrow T}$ and $P_{0 \rightarrow T}$ be the joints stochastic processes defined on the hypothesis class of the learning task, respectively. Define the true risk of a draw from $Q_{0 \rightarrow T}$  on an i.i.d. sample ${\bf Y} = \{ {\bf y}_{1}, \dots, {\bf y}_{K} \}$ at discrete and potentially irregular time points $t_1, \dots, t_K$ drawn from an unknown ground-truth stochastic process $\mathfrak{G}(t)$ as the expected model misfit as
 on the sample as defined via the following risk over hypotheses $H=({\bf h}_{0 \rightarrow T},\theta_f)$}
\begin{equation}
R(H) =  1 - \mdE_{{\bf Y} \sim \mathfrak{G}(t)} \Bigg [\prod_{k=1}^K p( {\bf y}_{k}|{\bf h}_{k})/\overline{B}   \Bigg ], \label{eqapp:true_risk}
\end{equation}
{\it for time horizon $T > 0$ and the corresponding empirical risk on a data set $\mathcal{D}=\{ {\bf Y}_1, \dots, {\bf Y}_N \}$ } \textit{as}
\begin{equation}
    R_\mcD(H) = 1-\frac{1}{N} \sum_{n=1}^N \Bigg [ \prod_{k=1}^K 
                p( {\bf y}_{k}^{n}|{\bf h}_{k})/\overline{B}  \Bigg ].
\end{equation}
{\it Then the expected true risk is bounded above by the marginal negative log-likelihood of the predictor and a complexity functional as}
 \begin{align}
  &\Ep{H\sim Q_{0 \rightarrow T}}{R(H)}  \leq \Ep{H\sim Q_{0 \rightarrow T}}{R_{\mathcal{D}}(H)} + \mathcal{C}_\delta(Q_{0 \rightarrow T},P_{0 \rightarrow T}), \label{eqapp:pac_bound}\\
         &~~~~\le - \frac{1}{N} \sum_{n=1}^N \log \left(\frac{1}{S} \sum_{s=1}^S \prod_{k=1}^K p({\bf y}_k^n|{\bf h}_{k}^{n,s})\right) + \mathcal{C}_{\delta/2}(Q_{0 \rightarrow T},P_{0 \rightarrow T})  +\sqrt{\frac{\log(2N/\delta)}{2S}} + K\log \overline{B}\\
         &~~~~\le -\frac{1}{SN}
         \sum_{n=1}^N \sum_{s=1}^S \sum_{k=1}^K \log \Big(p( {\bf y}_{k}^{n} |{\bf h}_{k}^{s,n}) \Big)   +  \mathcal{C}_{\delta/2}(Q_{0 \rightarrow T},P_{0 \rightarrow T})  +\sqrt{\frac{\log(2N/\delta)}{2S}} + K\log \overline{B}\label{eqapp:training_loss},
 \end{align}
{\it where $\overline{B} := \max_{{\bf y}_{k}, {{\bf h}_{k}}} p({\bf y}_{k}|{{\bf h}_{k}})$ is the uniform bound, $S$ is the sample count taken independently for each observed sequence, and the complexity functional is given as}
\begin{equation*}
		\mathcal{C}_\delta(Q_{0 \rightarrow T},P_{0 \rightarrow T}) := \sqrt{\frac{D_{KL}\big(Q_{0 \rightarrow T} || P_{0 \rightarrow T}\big) + \log({2\sqrt{N}}) - \log( {\delta/2})}{2N}}
\end{equation*}
{\it with $D_{KL}\big(Q_{0 \rightarrow T} || P_{0 \rightarrow T}\big)$ as in Lemma 1 for some $\delta > 0$.}

\paragraph{Proof.}
 To be able to apply known PAC bounds, we first define the hypothesis class $H \in \mathcal{H}_K$ that contain latent states $ {\bf{h}}_{k}, \theta_f$ that explain the observations $ {\bf y}_{k}$. Then, we define the true risk as  
 \begin{equation*}
    R(H) 
    =\mathds{E}_{{\bf{Y}}_{k} \sim \mathfrak{G}(t)}\Bigg[1 - \frac{1}{\overline{B}_K} \prod_{k=1}^K 
                p( {\bf y}_{k}|{\bf h}_{k}) \Bigg]
\end{equation*}
and the empirical risk as
\begin{equation*}
    R_\mcD(H)
   = \frac{1}{N} \sum_{n=1}^N
                \left\{ 1 - \frac{1}{\overline{B}_K} \prod_{kn=1}^K 
                p( {\bf y}_{k}^{n}|{\bf h}_{k}^{n}) \right\},
\end{equation*}
where we defined
\begin{equation*}
 \overline{B}_K := \max_{\mby,{\bf h}_{k}}\prod_{k=1}^Kp\left(\mby_k |{\bf h}_{k}\right) \le \left(\max_{\mby,{\bf h}_{k}}p\big(\mby_k|{\bf h}_{k}\big)\right)^K.
\end{equation*}
The data set $\mcD = \{{\bf{Y}}_{k}^{n}\}_{k,n}$ was generated by an unknown stochastic process $\mathfrak{G}(t)$. Note that we normalize the risks $R(H)$ and
$R_D(H)$ by the maximum of the likelihood and thereby obtaining a
possible range of these risk of $[0,1]$. The likelihood can be bounded, as the term $p( {\bf y}_{k}|{\bf h}_{k})$ can be bounded from above, as  we model this by a Gaussian.Therefore, it is bounded,  if we assume a minimal allowed variance.

To obtain a tractable bound, it is common practice is to upper bound its analytically
intractable inverse~\citep{germain2016pac} using Pinsker's
inequality~\citep{catoni2007pac, dziugaite2017computing}. 
Indeed, by applying Pinsker's inquality to the PAC-Theorem from \citet{maurer2004note}, we obtain the following theorem.

\paragraph{PAC-theorem}
For any $[0,1]$-valued loss function giving rise to empirical and true risk $R_\mcD(H),R(H)$,  for any distribution $\Delta$, for any $N\in \mathds{N}$, $N > 8$, for any distribution $P_{0 \rightarrow T}$ on a hypothesis set $\mathcal{Q}_K$, and for any $\delta \in (0,1]$, the following holds with probability at least $1 - \delta$ over the training set $\mcD\sim \Delta^N$:
\begin{equation*}
  \forall Q_{0 \rightarrow T}: \quad \Ep{H\sim Q_{0 \rightarrow T}}{R(H)} \le \Ep{H\sim Q_{0 \rightarrow T}}{R_\mcD(H)} + \sqrt{\frac{\KL{ Q_{0 \rightarrow T}}{P_{0 \rightarrow T}} + \log\left(\frac{2\sqrt{N}}{\delta}\right)}{2N}}
\end{equation*}

Here, $\KL{ Q_{0 \rightarrow T}}{P_{0 \rightarrow T}}$ acts as a complexity measure that measures, how
much the posterior predictive governing the SDE $ Q_{0 \rightarrow T}$ needed to be adapted to the data when compared to an a priori chosen SDE that could alternatively have generated data $P_{0 \rightarrow T}$. In
our situation, $Q_{0 \rightarrow T}$ is obtained by our approximation scheme, resulting in a bounded likelihood of observations ${\bf y}_{k}$ which factorizes
over different observations $n$. The $P_{0 \rightarrow T}$ can be arbitrarily
chosen as long as it does not depend on the observations. As mentioned in the main paper, we chose an SDE with the same diffusion term which also factorizes over observations. Using this setting, we can analytically compute the KL-distance (as shown in  Lemma~1).

On the right hand side of this PAC-bound, we need to
evaluate $\Ep{H\sim Q_{0 \rightarrow T}}{R_\mcD(H)}$. To this end, we note
\begin{align*}
      \Ep{{H}\sim Q_{0 \rightarrow T}}{R_\mcD(H)} &=  \frac{1}{N}  \sum_{n=1}^N  \mathds{E}_{H\sim Q_{0 \rightarrow T}}
                \left[ 1 - \frac{1}{\overline{B}_K} \left(\prod_{k=1}^K 
                p( {\bf y}_{k}^{n}|{\bf h}_{k}^{n}) \right) \right]
\\
&=1-\frac{1}{N}\sum_{n=1}^N\mathds{E}_{H\sim Q_{0 \rightarrow T}}\left[ \frac{1}{\overline{B}_K}\prod_{k=1}^K
                p( {\bf y}_{k}^{n}|{\bf h}_{k}^{n})\right]\\
& \stackrel{\text{\tiny Hoeffding}}{\le} 1-\frac{1}{SN}\sum_{n=1}^N\sum_{s=1}^S\Bigg[ \frac{1}{\overline{B}_K}\prod_{k=1}^K 
                p( {\bf y}_{k}^{n}|{\bf h}_{k}^{n,s})  \Bigg] + \sqrt{\frac{\log(2N/\delta)}{2S}}\\
                &= \frac{1}{N}\sum_{n=1}^N\left\{1-\frac{1}{S}\sum_{s=1}^S\Bigg[ \frac{1}{\overline{B}_K}\prod_{k=1}^K 
                p( {\bf y}_{k}^{n}|{\bf h}_{k}^{n,s})  \Bigg]\right\} + \sqrt{\frac{\log(2N/\delta)}{2S}}\\
    &\stackrel{-\log(z)\geq 1-z}{\leq}   -\frac{1}{N}\sum_{n=1}^N \log\left(\frac{1}{S}\sum_{s=1}^S \prod_{k=1}^K p( {\bf y}_{k}^{n}|{\bf h}_{k}^{n,s})  \right)+ \log {\overline{B}_K}+ \sqrt{\frac{\log(2N/\delta)}{2S}}\\
&\stackrel{\text{\tiny{Jensen's ineq.}}}{\leq}-\frac{1}{SN}\sum_{n=1}^N\sum_{s=1}^S \sum_{k=1}^{K}\big[\log 
                p( {\bf y}_{k}^{n}|{\bf h}_{k}^{n,s})  \big]+ \log {\overline{B}_K}+ \sqrt{\frac{\log(2N/\delta)}{2S}},
\end{align*}
where we have used Hoeffding's inequality for estimating the true expectation over hypotheses with a $K$ samples trace  $ {\bf h}_{k}^{n,s}, k=1,\dots,K, s=1,\dots,S$ for each observation $n$. As we approximate the integral for each time-series $n$ separately via sampling, we require Hoeffding to hold simultaneously for all $n$. Using a union bound, we have to scale $\delta$ for each $n$ by $N$. Splitting confidences between the PAC-bound and the sampling based approximation results an additional factor of 2. With  $\delta/(2N)$, the corresponding inequality holds with a probability of $\mdP> \delta/2$. Also using $\delta/2$ in PAC-theorem, we obtain that with $\mdP \geq 1 - \delta$ we have for all $Q_{0 \rightarrow T}$ that
\begin{align*}
   &\Ep{H\sim Q_{0 \rightarrow T}}{R(H)} \le \Ep{H\sim Q_{0 \rightarrow T}}{R_\mcD(H)} + \sqrt{\frac{\KL{ Q_{0 \rightarrow T}}{P_{0 \rightarrow T}} + \log\left(\frac{2\sqrt{N}}{\delta/2}\right)}{2N}}\\
&\leq-\frac{1}{N}\sum_{n=1}^N \log\left(\frac{1}{S}\sum_{s=1}^S \prod_{k=1}^K p( {\bf y}_{k}^{n}|{\bf h}_{k}^{n,s})  \right) + \sqrt{\frac{\KL{ Q_{0 \rightarrow T}}{P_{0 \rightarrow T}} + \log\left(\frac{2\sqrt{N}}{\delta/2}\right)}{2N}} + \log {\overline{B}_K}+ \sqrt{\frac{\log(2N/\delta)}{2S}}\\
&\leq -\frac{1}{SN}\sum_{n=1}^N\sum_{s=1}^S \sum_{k=1}^{K}\big[\log 
                p( {\bf y}_{k}^{n}|{\bf h}_{k}^{n,s})  \big]+ \sqrt{\frac{\KL{ Q_{0 \rightarrow T}}{P_{0 \rightarrow T}} + \log\left(\frac{2\sqrt{N}}{\delta/2}\right)}{2N}} + \log {\overline{B}_K}+ \sqrt{\frac{\log(2N/\delta)}{2S}}
\end{align*}$\hfill \qed$

\paragraph{Corollary 1.} {\it Given a $L$-Lipschitz continuous function set}
 \begin{align*}
    \Big \{f_{\theta}^{n}(x)&:\mdR \rightarrow [0,1] \Big | n=1,\cdots,N \Big \} \bigcup\Big \{ g_{\theta}(x): \mdR \rightarrow [0,+\infty] \Big \},
 \end{align*}
 {\it for the two losses:}
 \begin{align*}
     l_1(\theta)  &= -\sum_{n=1}^N f_{\theta}^n(x)  + g_{\theta}(x)\quad\textit{and}\quad l_2(\theta)  =  -\sum_{n=1}^N \log f_{\theta}^n(x)  + g_{\theta}(x),
 \end{align*}
{\it the sequential updates ($\theta^{0} := \theta$)}
\begin{align*}
    \theta^{(n)} &\leftarrow \theta^{(n-1)} + \alpha_{n} \nabla \big(\log f_{\theta^{(n-1)}}^n(x)\big),~n = 1,\ldots,N,\\
     \theta^{(N+1)} &\leftarrow \theta^{(N)} - \alpha_{N+1} \nabla g_{\theta^{(N)}}(x),
\end{align*}
{\it where $\alpha_n \in (0, f_{\theta^{(n-1)}}^n(x)/L)~\forall n$ and $\alpha_{N+1} \in (0, 1/L)$, satisfy both $l_1(\theta^{(N+1)}) \leq l_1(\theta)$ and $l_2(\theta^{(N+1)}) \leq l_2(\theta)$.}

\paragraph{Proof.} As we only consider updates in $\theta$ for constant $x$, we simplify the notation for this proof to $f^{n}(\theta) := f_\theta^{n}(x)$, $g(\theta) = g_\theta(x)$. I.e.\ we have as the two loss terms 
\begin{align*}
    l_1(\theta) &= -\sum_{n=1}^N f^n(\theta) + g(\theta)\quad \text{and}\quad l_2(\theta) = -\sum_{n=1}^N \log f^n(\theta) + g(\theta).
\end{align*}
In general we have with $\log f(\theta) < f(\theta)$ that $l_1(\theta) < l_2(\theta)$. Similarly we have 
\begin{align*}
    \nabla l_2(\theta) &= - \sum_n \underbrace{\frac{1}{f^n(\theta)}}_{\geq 1} \nabla f^n(\theta) + \nabla g(\theta) \leq -\sum_n \nabla f^n(\theta) + \nabla g(\theta)  = \nabla l_1(\theta).
\end{align*}
Due to the sequential updates we can consider each term separately. For an $L$-Lipschitz function $f^n(\theta)$, we have that for arbitrary $x,y$
\begin{equation*}
    f(y) \leq f(x) + \nabla f(x)^\top(y - x) + \frac L2||y - x||_2^2.
\end{equation*}
Choosing $y = \theta^{(n-1)}$ and $x = \theta^{(n)} = \theta^{(n-1)} + \alpha_n \nabla \log f^n$ this gives us 
\begin{align*}
    f(\theta^{(n-1)}) &\leq f(\theta^{(n)}) - \frac{\alpha_n}{f^n(\theta^{(n)})}||\nabla f^n(\theta^{(n)})||_2^2 + \frac{L\alpha_n^2}{2f^n(\theta^{(n)})^2}||\nabla f^n(\theta^{(n)})||_2^2\\
    &= f^n(\theta^{(n)}) - \underbrace{\frac{\alpha_n}{f^n(\theta^{(n)})}}_{\geq 0}\underbrace{\left(1 - \frac{L\alpha_n}{2f^n(\theta^{(n)})}\right)}_{> 0}||\nabla f^n(\theta^{(n)})||_2^2\leq f^n(\theta^{(n)}),
\end{align*}
and hence chaining the update steps gives the desired result. \qed

That is, updating the terms in $l_2(\theta)$ sequentially, one can ensure concurrent optimization of $l_1(\theta)$. Note that $l_1(\theta)$ and $l_2(\theta)$ are not necessarily dual objectives, hence may have different extrema. Nevertheless, a gradient step that decreases one loss also decreases the other with potentially a different magnitude. In practice, we observe this behavior to also hold empirically for joint gradient update steps with shared learning rates. Applying Lemma 2 to the setup in Theorem 2, we establish a useful link between Empirical Bayes and PAC learning.

\paragraph{Theorem 2 (strong convergence).} {\it Let ${\bf h}_t^{\theta}$ be an It\^{o} process as in 
(4)
with drift parameters $\theta$  and its Euler-Maruyama approximation $\widetilde{\bf h}_t^{\theta}$ for some regular step size $\Delta t>0$. For some coefficient $R>0$ and any $T>0$, the following inequality holds }
\begin{equation*}
    \mathds{E} \Bigg [ \sup_{0 \leq t \leq T} \Big | \mathds{E}_{\theta}[{\bf h}_t^{\theta}] - \dfrac{1}{S} \sum_{s=1}^S \widetilde{\bf h}_t^{\theta^{(s)}} \Big | \Bigg ] \leq R \Delta t^{1/2},
\end{equation*}
{\it as $S\to \infty$, where $\{\theta^{(s)}\sim p_{\phi}(\theta_f)|s=1,\ldots,S\}$ are i.i.d.\ draws from a prior $p_{\phi}(\theta_f)$. }

\paragraph{Proof: } The Euler-Maruyama (EM) approximation converges strongly as
\begin{equation*}
    \mdE\left[\big|\mbh_T^\theta - \widetilde \mbh_T^\theta\big|\right] \leq R\Delta t^{1/2},
\end{equation*}
for a positive constant $R$ and a suitably small step size $\Delta t$ as discussed e.g.\ by~\citet{kloeden2011numerical}. To simplify the mathematical notation we follow their approach of comparing the absolute error of the end of the trajectory throughout the proof. 
As our sampling scheme is unbiased it is a consistent estimator and we have that asymptotically for $S\to \infty$
\begin{equation*}
    \frac1S\sum_{s=1}^S\widetilde h_T^{\theta^(s)}  = \mdE_\theta [\widetilde h_T^\theta].
\end{equation*}
We then have for the marginal $\mbh_T$, $\tilde \mbh_T$ that 
\begin{align*}
    \mdE\left[\big|\mbh_T - \widetilde\mbh_T\big|\right] &=\mdE\left[\big|\mdE_\theta\mbh_T^\theta - \mdE_\theta\widetilde\mbh_T^\theta\big|\right]\\
    &= \mdE\left[\big|\mdE_\theta\left[\mbh_T^\theta - \widetilde\mbh_T^\theta\right]\big|\right]\\
    &\leq \mdE\left[\mdE_\theta\left[\big|\mbh_T^\theta - \widetilde\mbh_T^\theta\big|\right]\right]\\
    &\leq \mdE_\theta\left[R\Delta t^{1/2}\right] = R\Delta t^{1/2},
\end{align*}
where the first inequality is due to Jensen and the second due to the strong convergence result for a fixed set of parameters. \qed
\section{COMPUTATIONAL COST}
We present the runtimes of the different approaches in Table~\ref{tab:runtime}.
D-BNN samples the weights of the neural network directly leading to the runtime term $\mathcal{O}(MTF)$. 
All other approches do not sample the weights but the linear activations of the each data points leading to $\mathcal{O}(2MTF)$.
When we apply empirical Bayes, we dot not use any regularization term on the weights, while all other approaches contain a penalty term with cost $\mathcal{O}(W)$. 
Using the PAC-framework, we employ a second regularization term that leads to an additional runtime cost of $\mathcal{O}(TMD^3)$. 
However, the cubic cost in $D$ is invoked by inverting the diffusion matrix $G (h_t, t)$ and can be further reduced by choosing a simpler form for $G(h_t, t)$ (e.g.\ diagonal). 
In case that prior knowledge is available in ODE form, we need to compute the corresponding drift term for each time point and each MC sample leading to the term $\mathcal{O}(MTP)$.
\begin{table*}[h]
	\caption{ Computational cost analysis in FLOPs for time series of length \textbf{T}.
	\textbf{M:} Number of Monte Carlo Samples. \textbf{W:} Number of weights in the neural net.
	\textbf{F:} Forward pass cost of a neural net. \textbf{L:} Cost for computing the likelihood term.
	\textbf{D:} Number of dimensions. \textbf{P:} Cost of a prior SDE integration.
	}
	\begin{center}
			\begin{tabular}{lc}
		   				\toprule
			\textbf{Model} &  \textbf{Training per Iteration}  \\
				\midrule
				D-BNN (SGLD) & $\mathcal{O}(MTF + MTDL + W)$ \\
				Variational Bayes & $\mathcal{O}(2MTF + MTDL + W)$  \\
				$\text{E-Bayes}$       & $\mathcal{O}(2MTF + MTDL)$ \\
				$\text{E-PAC-Bayes}$ & $\mathcal{O}(2MTF + MTDL + W + TMD^3)$ \\
				$\text{E-Bayes-Hybrid}$  & $\mathcal{O}(2MTF+ MTDL + MTP)$  \\
				$\text{E-PAC-Bayes-Hybrid}$   & $\mathcal{O}(2MTF + MTDL + W + TMD^3 + MTP)$  \\
				\bottomrule
			\end{tabular}
	\end{center}
	\label{tab:runtime}
\end{table*}
\section{FURTHER DETAILS ON THE EXPERIMENTS}\label{app:expdetails}

Here we provide the details of the experiment setup we used in obtaining our results reported in the main paper. We observed our results to be robust against most of the design choices. We provide a reference implementation at \url{https://github.com/manuelhaussmann/bnsde}.

\subsection{Lorenz Attractor}

We took $200000$ Euler-Maruyama steps ahead with a time step size of $10^{-4}$ and downsampling by factor $0.01$, which gives a sequence of $2000$ observations with frequency $0.01$. We split the first half of this data set into 20 sequences of length 50 and use them for training, and the second half to 10 sequences of length 100 and use for test. For all model variants, we used an Adam optimizer learning rate $0.001$, minibatch size of two, a drift net with two hidden layers of $100$ neurons and softplus activation function.%
We trained all models for $100$ epochs and observed this training period to be sufficient for convergence. 

\subsection{CMU Motion Capture}

In this experiment, we tightly follow the design choices reported by \citet{yildiz19odevae} to maintain commensurateness. This setup assumes the stochastic dynamics are determined in a six-dimensional latent space. \citet{yildiz19odevae} use an auto-encoder to map this latent space to the $50-$dimensional observation space back and forth. We adopt their exact encoder-decoder architecture and incorporate it into our BNSDE, arriving at the data generating process 
\begin{align*}
    \theta_f &\sim p_{\phi_f}(\theta_f),\\
    d{\bf h}_t|\theta_f &\sim  f_{ \theta_f }\big( b_{\lambda}({\bf h}_t),t\big) dt + G\big(b_{\lambda}({\bf h}_t),t\big) d\boldsymbol{\beta}_t, \\
    {\bf z}_t | {\bf h}_t &\sim \mathcal{N}({\bf z}_t | a_{\psi}({\bf h}_t), 0.5\cdot10^{-6} \mathds{1}), \\
    {\bf y}_{t} | {\bf z}_{t} &\sim \mathcal{N}(\mby_{t}|{\bf z}_{t},0.5\cdot10^{-6} \mathds{1}),\qquad\forall t \in {\bf t}.
\end{align*}
Above, $b_{\lambda}(\cdot,\cdot)$ is the encoder which takes the observations of the last three time points as input, passes them through two dense layers with 30 neurons and softplus activation function, and then linearly projects them to a six-dimensional latent space, where the dynamics are modeled. The decoder $a_{\psi}({\bf h}_t)$ follows the same chain of mapping operations in reverse order. The only difference is that the output layer of the decoder emits only one observation point, as opposed to the encoder admitting three points at once. 

The drift function $f_{ \theta_f }(\cdot,\cdot)$ is governed by another separate Bayesian neural net, again with one hidden layer of 30 neurons and softplus activation function on the hidden layer.
The diffusion function is fixed to be a constant.

We train all models except SGLD with the Adam optimizer for $3000$ epochs on seven randomly chosen snippets at a time with a learning rate of $10^{-3}$. We use snippet length 30 for the first 1000 epochs, 50 until epoch 2500, and 100 afterwards. SGLD demonstrates significant training instability for this learning rate, hence for it we drop its learning rate to the largest possible stable value $10^{-5}$ and increase the epoch count to 5000.

\section{FURTHER EXPERIMENTS}
\subsection{Lotka Volterra}
\rtodo{Update text to ensure consistency now that it is in the appendix}
\rtodo{Update legend in the figure with the final method names!}
\begin{figure}
    \centering
    \includegraphics[width=0.6\linewidth]{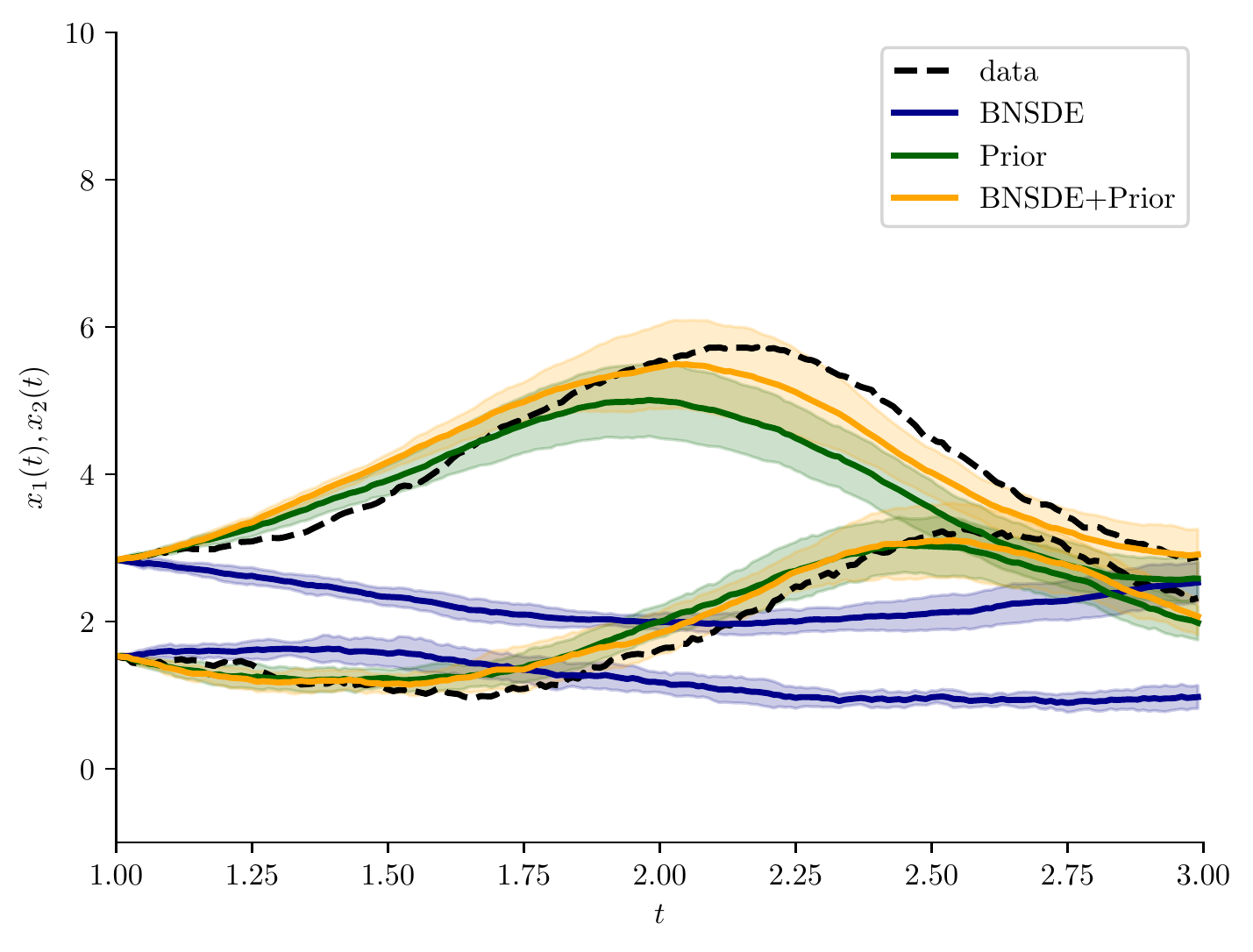}
    \caption{Lotka-Volterra visualization. Error bars indicate three standard deviations over 10 trajectories starting from the true value at $t=1$. The predictions over 200 time steps ($dt=0.01$) are for: \textit{i)} a BNSDE trained without prior knowledge, \textit{ii)} an SDE with known prior parameters, \textit{iii)} the joint hybrid BNSDE. The dashed lines are the observed trajectories for $x_t$ and $y_t$.}
    \label{fig:explotvol}
\end{figure}
We  demonstrate the benefits of incorporating prior knowledge although it is a coarse approximation to the true system. We consider the Lotka-Volterra system specified as: 
\begin{align*}
    dx_t &=  (\theta_1 x_t -  \theta_2 x_{t} y_{t})dt + 0.2~ d\beta_t,\\
    dy_t &= (-\theta_3 y_t + \theta_4 x_{t} y_{t})dt + 0.3 ~d\beta_t.
\end{align*}
with $\mbtheta = (2.0, 1.0, 4.0, 1.0)$. 
Assuming that the trajectory is observed on the interval $t=[0,1]$ with a resolution of $dt = 0.01$, we compare the following three methods: \textit{i)} the black-box BNSDE without prior knowledge, \textit{ii)} the white-box SDE in
(7)
representing partial prior knowledge (parameters are sampled from a normal distribution centered on the true values with a standard deviation of~$0.5$), and finally \textit{iii)} combining them in our proposed hybrid method. The outcome is summarized in Figure~\ref{fig:explotvol}. While the plain black-box model delivers a poor fit to data, our hybrid BNSDE brings significant improvement from relevant but inaccurate prior knowledge. 

\subsubsection{Experimental details}
We took $10^5$ Euler-Maruyama steps on the interval $[0,10]$ with a time step size of $10^{-4}$, downsampling them by a factor of $100$ giving us $1000$ observations with a frequency of $0.01$. We take the first $500$ observations on the interval [0,5] to be the training data and the observations in $(5,10]$ to be the test data. Each sequence is split into ten sequences of length $50$. Assuming the diffusion parameters to be known and fixed, both BNSDEs (i.e.\ with and without prior knowledge) get a 4 layer net as the drift function with $50$ neurons per layer and ReLU activation functions.
The BNSDE with prior knowledge as well as the raw SDE estimate each get an initial sample of $\tilde \mbtheta$ parameters as the prior information by sampling from a normal distribution centered around the true parameters ($\tilde \mbtheta \sim \Norm(\tilde \mbtheta|\mbtheta, \sigma^2\mathds{1}_4)$).
 The models are each trained for $50$ epochs with the Adam optimizer and a learning rate of $1e-3$. Since both the latent and observed spaces are only two dimensional, we did not need an observation model in this experiment. We directly linked the BNSDE to the likelihood.

\subsection{Lorenz Attractor}
\begin{figure}
    \centering
    \includegraphics[width=0.7\linewidth]{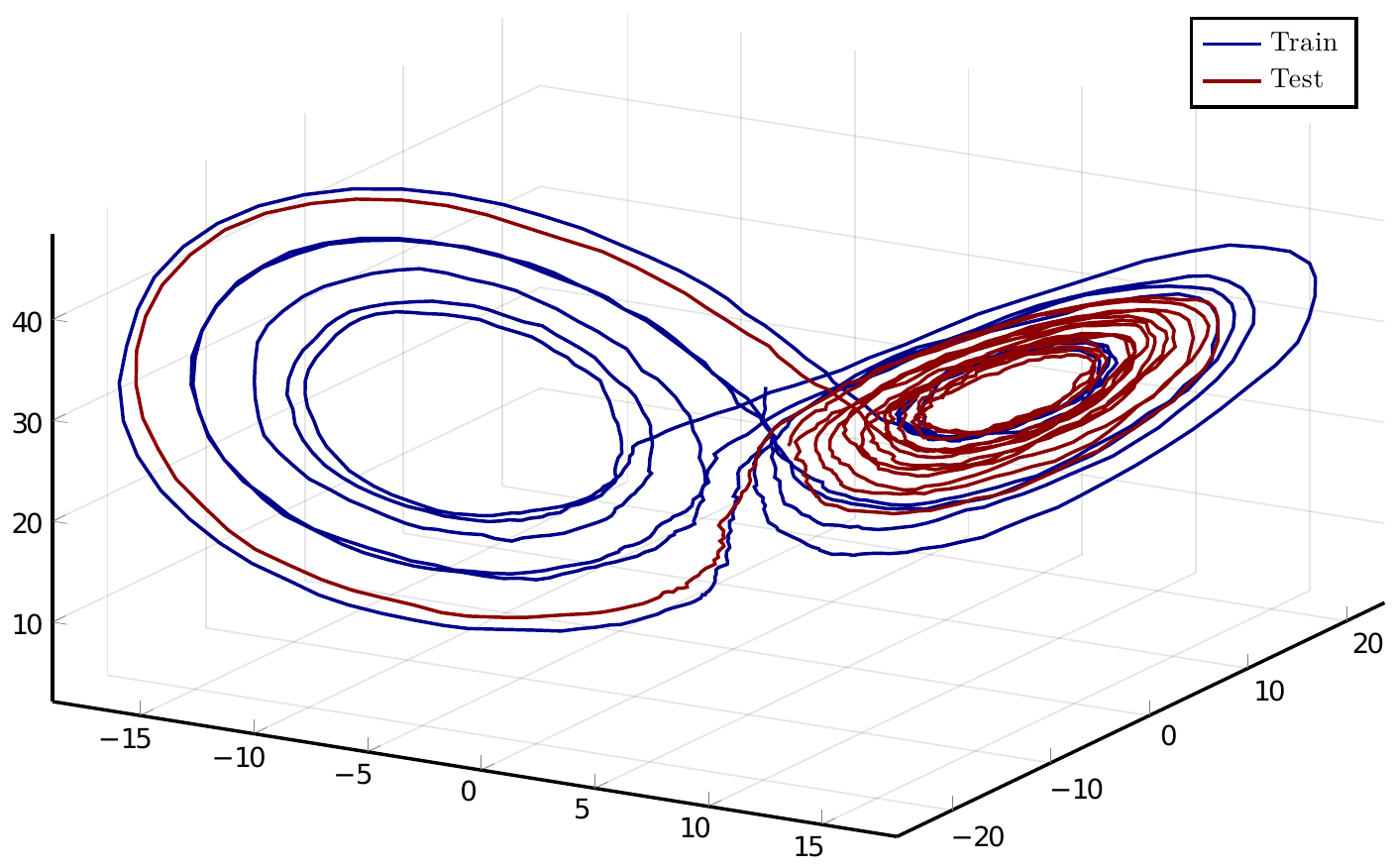}
    \caption{Visualization of the stochastic Lorenz attractor. Of the 2000 observations, the first 1000 constitute the training data (marked in blue), while the second 1000 are the test observations (marked in red). Note the qualitative difference of the two sets.}
    \label{fig:lorenz}
\end{figure}
As discussed in the main paper, the model is trained solely on the first 1000 observations of a trajectory consisting of 2000 observations, leaving the second half for the test evaluation. Figure~\ref{fig:lorenz} visualizes the qualitative difference between the two. Note also the single loop the trajectory performs which we will see again in the 1d projections below. To visualize explore the qualitative difference of our proposed model with weak prior knowledge compared to one lacking this knowledge we consider the situation where we we have structural prior knowledge only about the third SDE (i.e.\ the penultimate case in Table~1 with $\mbrho = [0,0,1]$.

\begin{figure} 
\centering 
\begin{subfigure}{0.8\linewidth}
\centering 
\includegraphics[width=\linewidth]{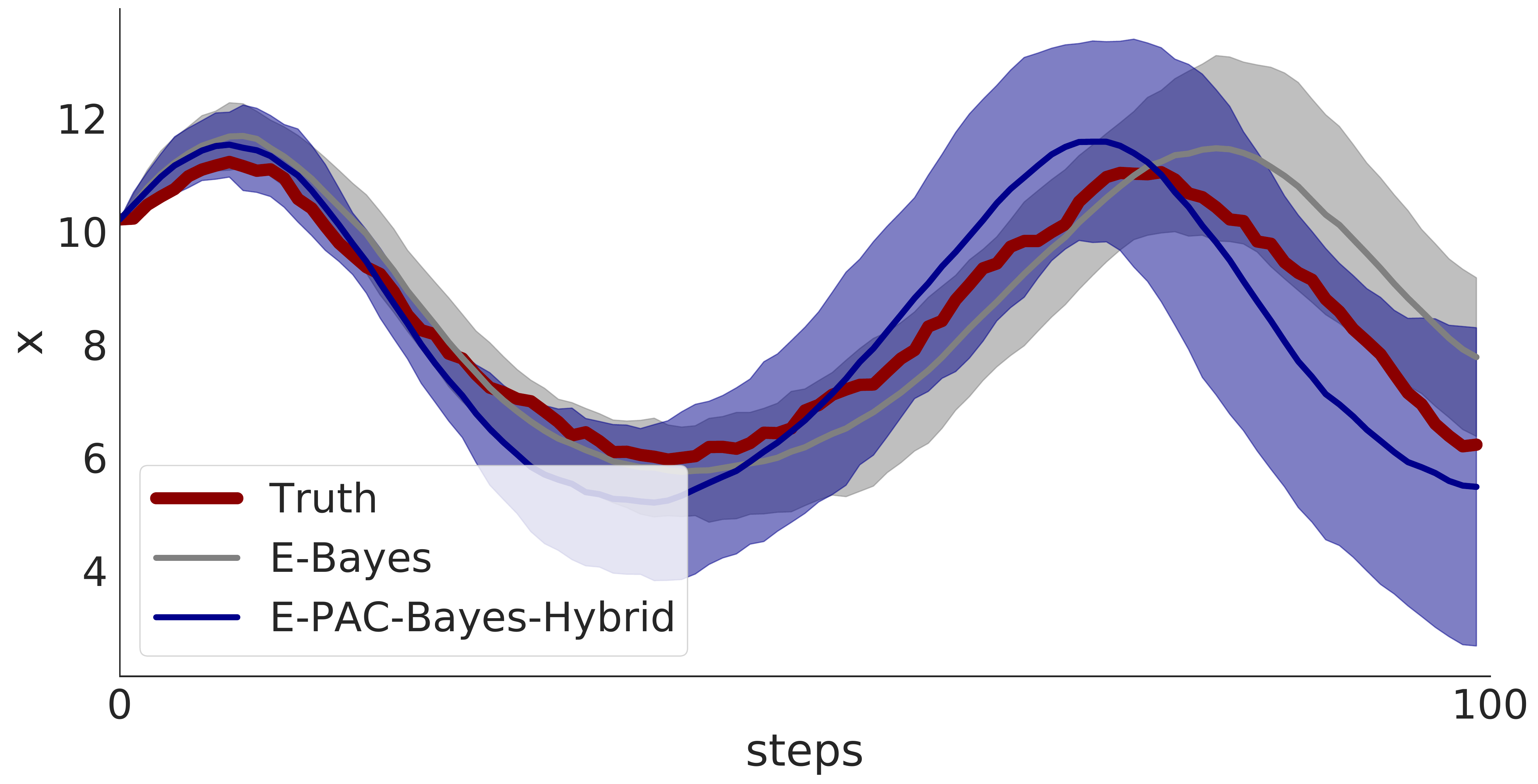}
\caption{$x$ coordinate over time}
\end{subfigure}
\begin{subfigure}{0.8\linewidth}
\centering 
\includegraphics[width=\linewidth]{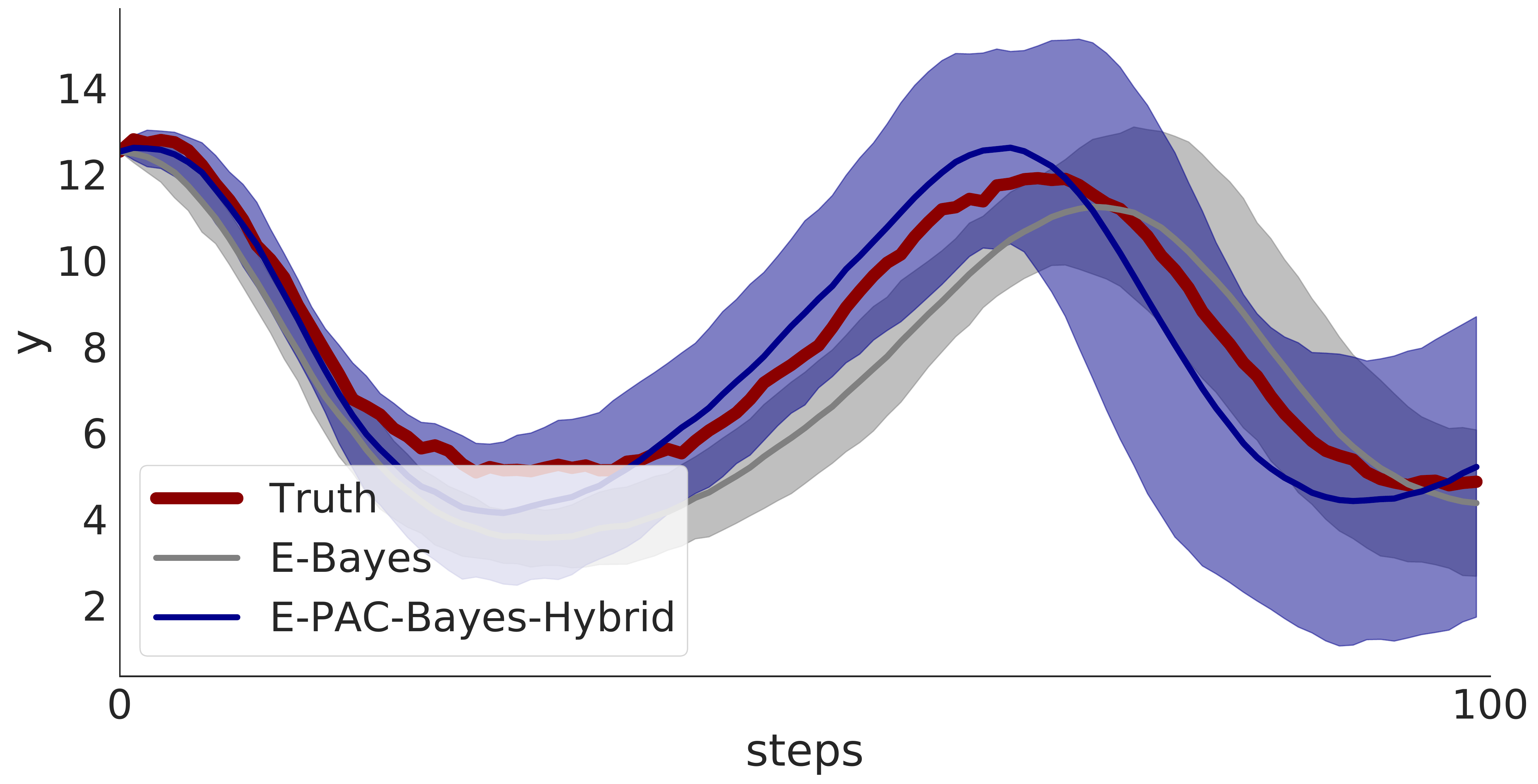}
\caption{$y$ coordinate over time}
\end{subfigure}
\begin{subfigure}{0.8\linewidth}
\centering 
\includegraphics[width=\linewidth]{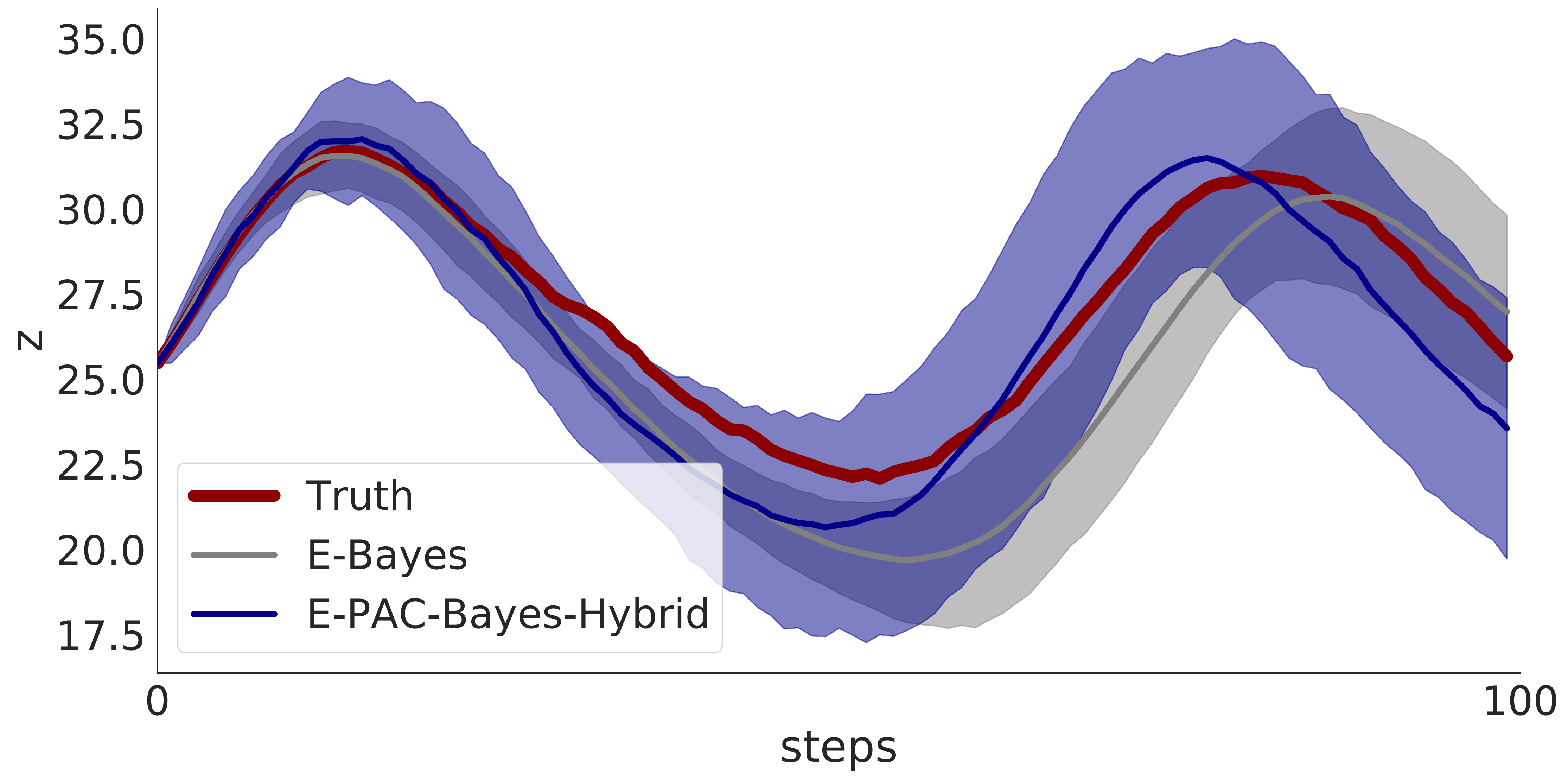}
\caption{$z$ coordinate over time}
\end{subfigure}
\caption{Predicting 100 time steps ahead.}
\label{fig:lorenz100}
\end{figure}

\begin{figure} 
\centering 
\begin{subfigure}{0.8\linewidth}
\centering 
\includegraphics[width=\linewidth]{figures/trajectories/200dtsteps-0dim.pdf}
\caption{$x$ coordinate over time}
\end{subfigure}
\begin{subfigure}{0.8\linewidth}
\centering 
\includegraphics[width=\linewidth]{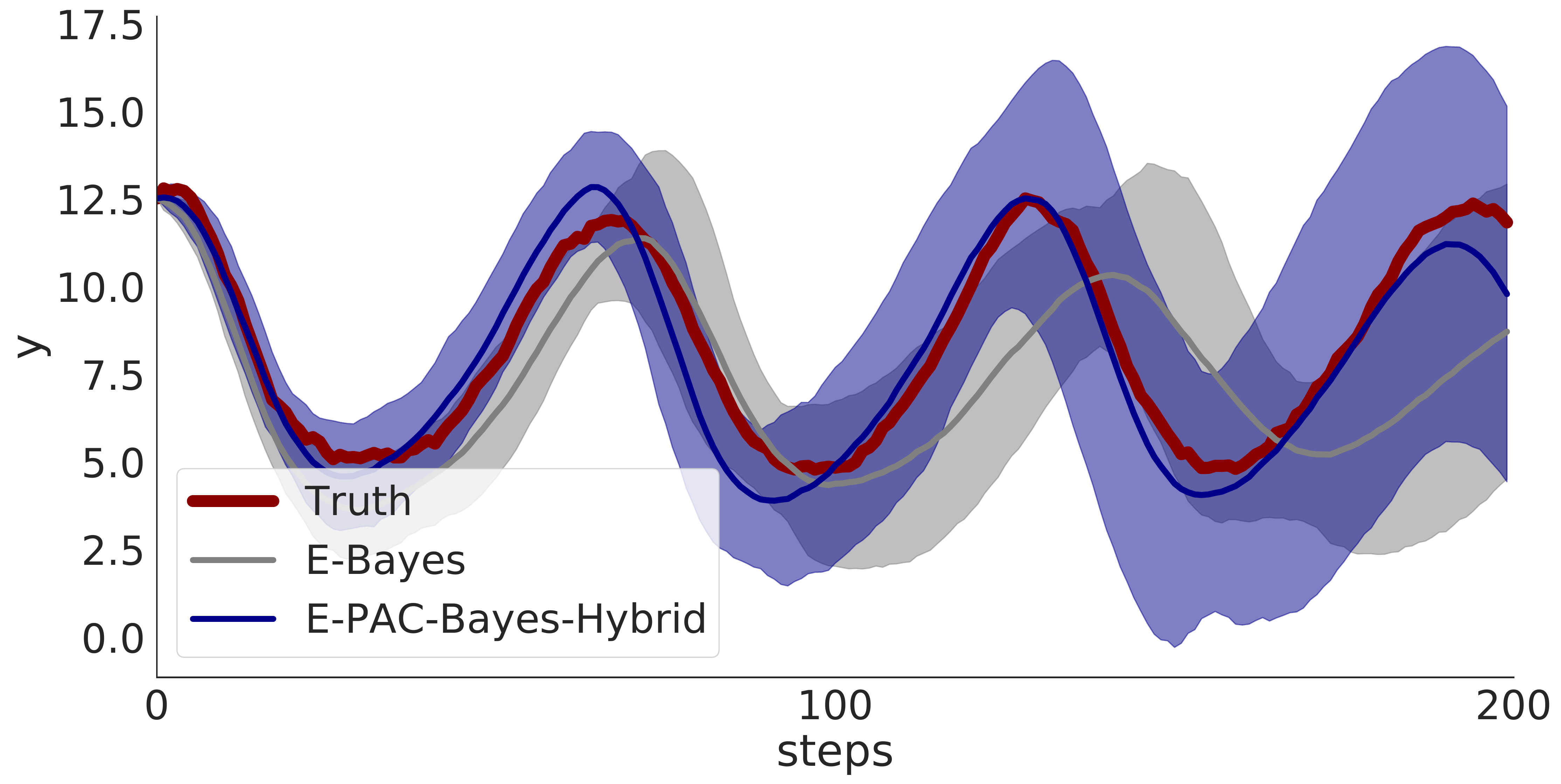}
\caption{$y$ coordinate over time}
\end{subfigure}
\begin{subfigure}{0.8\linewidth}
\centering 
\includegraphics[width=\linewidth]{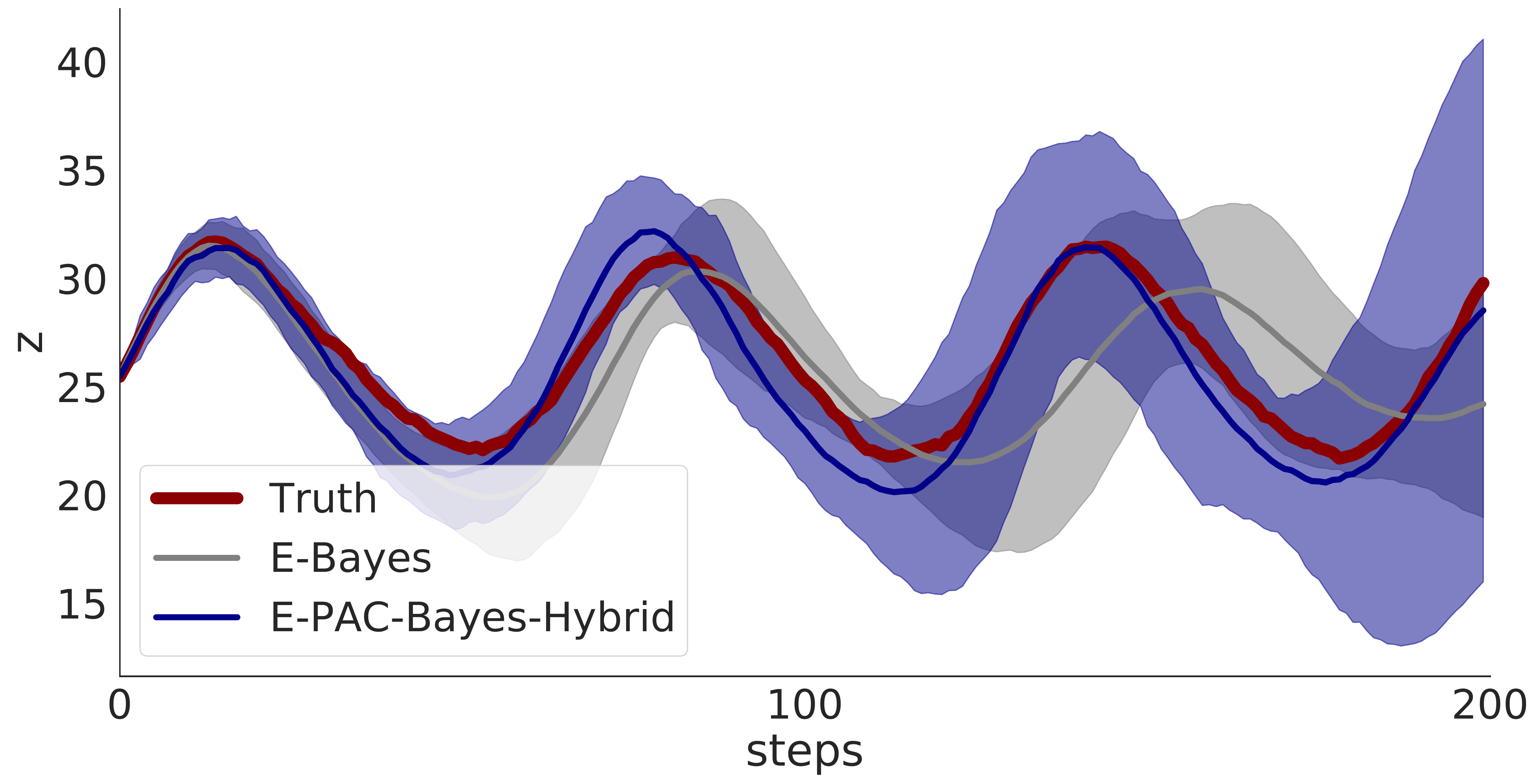}
\caption{$z$ coordinate over time}
\end{subfigure}
\caption{Predicting 200 time steps ahead.}
\label{fig:lorenz200}
\end{figure}

\begin{figure} 
\centering 
\begin{subfigure}{0.8\linewidth}
\centering 
\includegraphics[width=\linewidth]{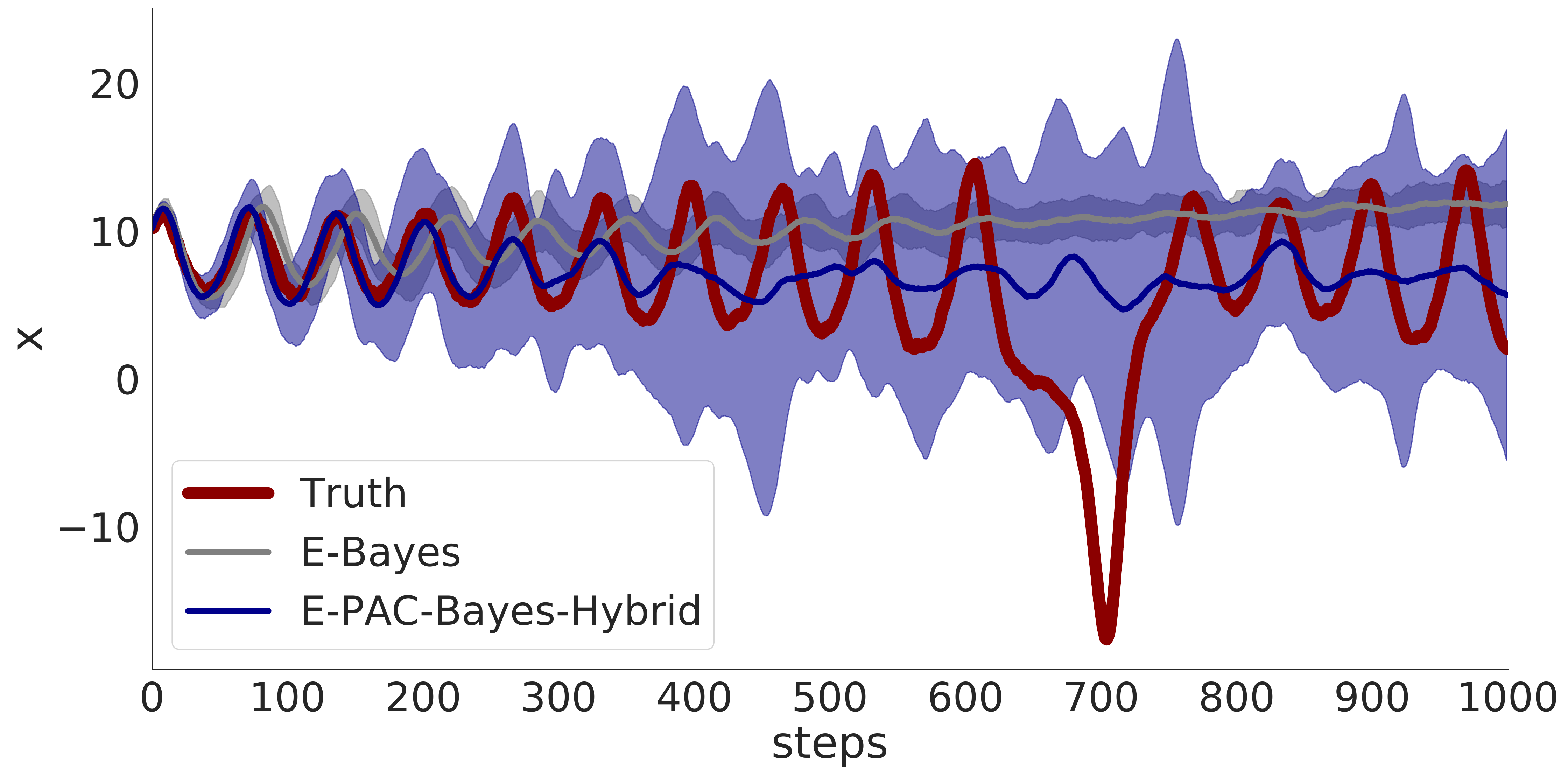}
\caption{$x$ coordinate over time}
\end{subfigure}
\begin{subfigure}{0.8\linewidth}
\centering 
\includegraphics[width=\linewidth]{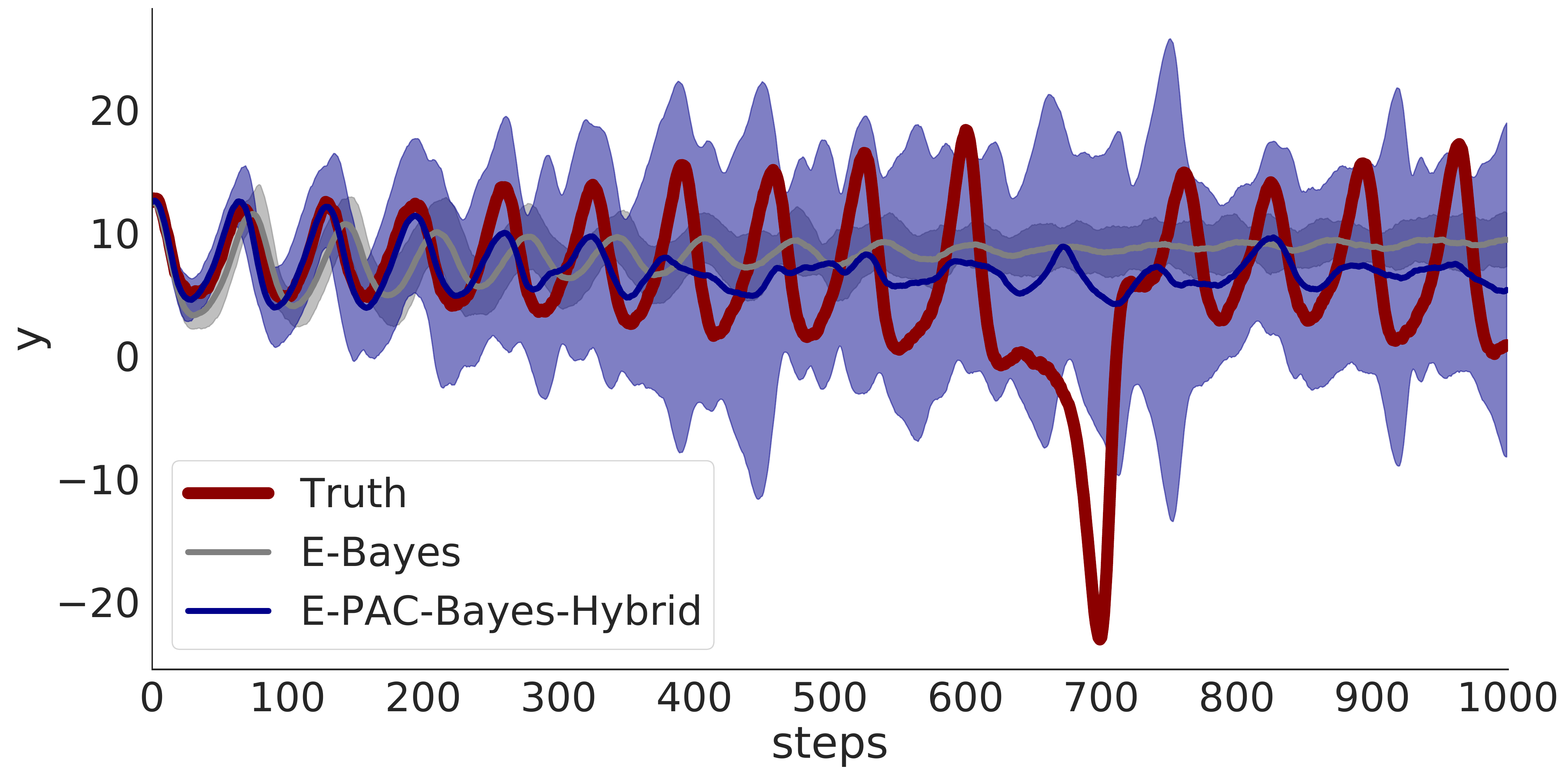}
\caption{$y$ coordinate over time}
\end{subfigure}
\begin{subfigure}{0.8\linewidth}
\centering 
\includegraphics[width=\linewidth]{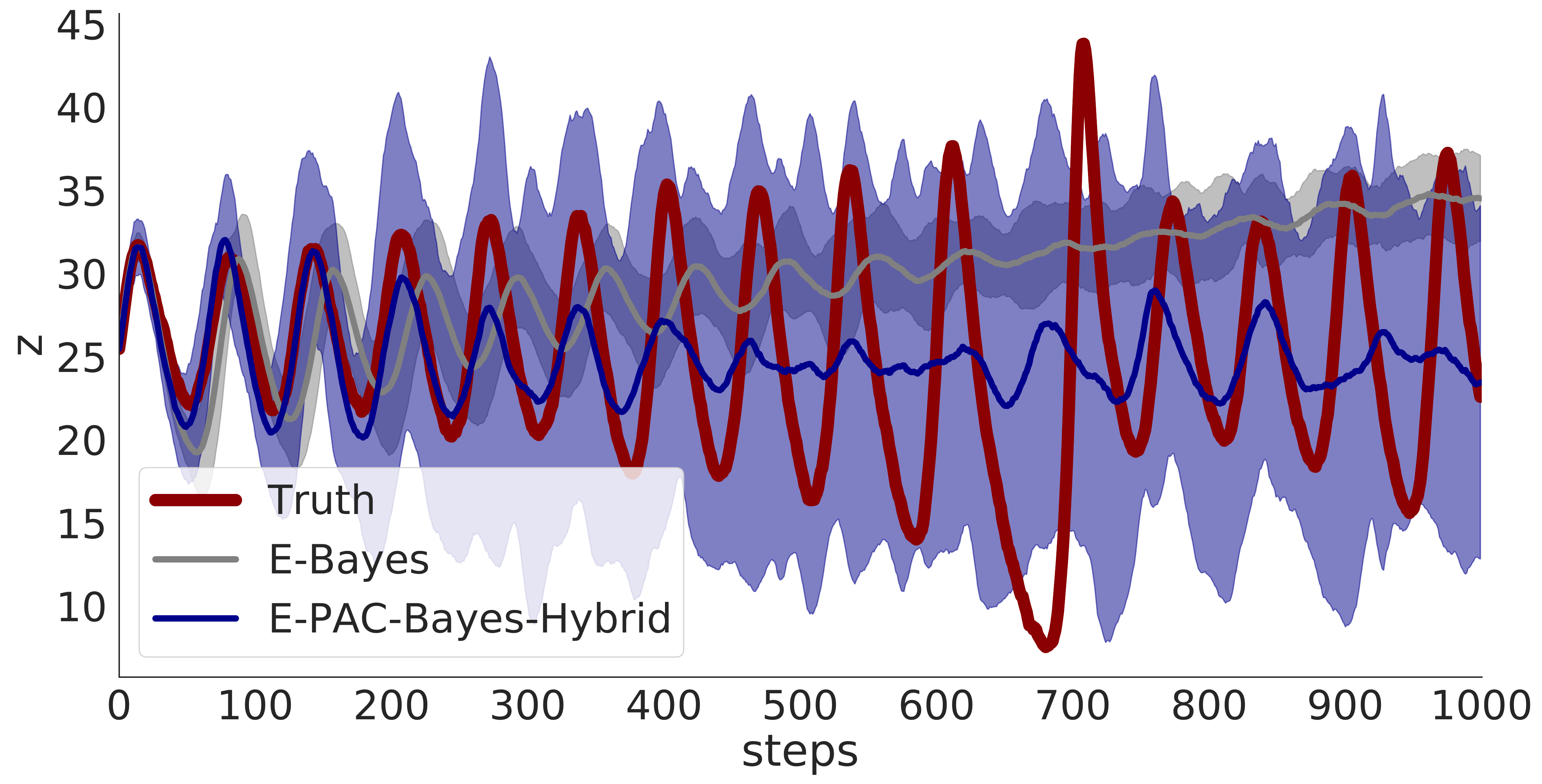}
\caption{$z$ coordinate over time}
\end{subfigure}
\caption{Predicting 1000 time steps ahead.}
\label{fig:lorenz1000}
\end{figure}

\begin{figure} 
\centering 
\begin{subfigure}{0.8\linewidth}
\centering 
\includegraphics[width=\linewidth]{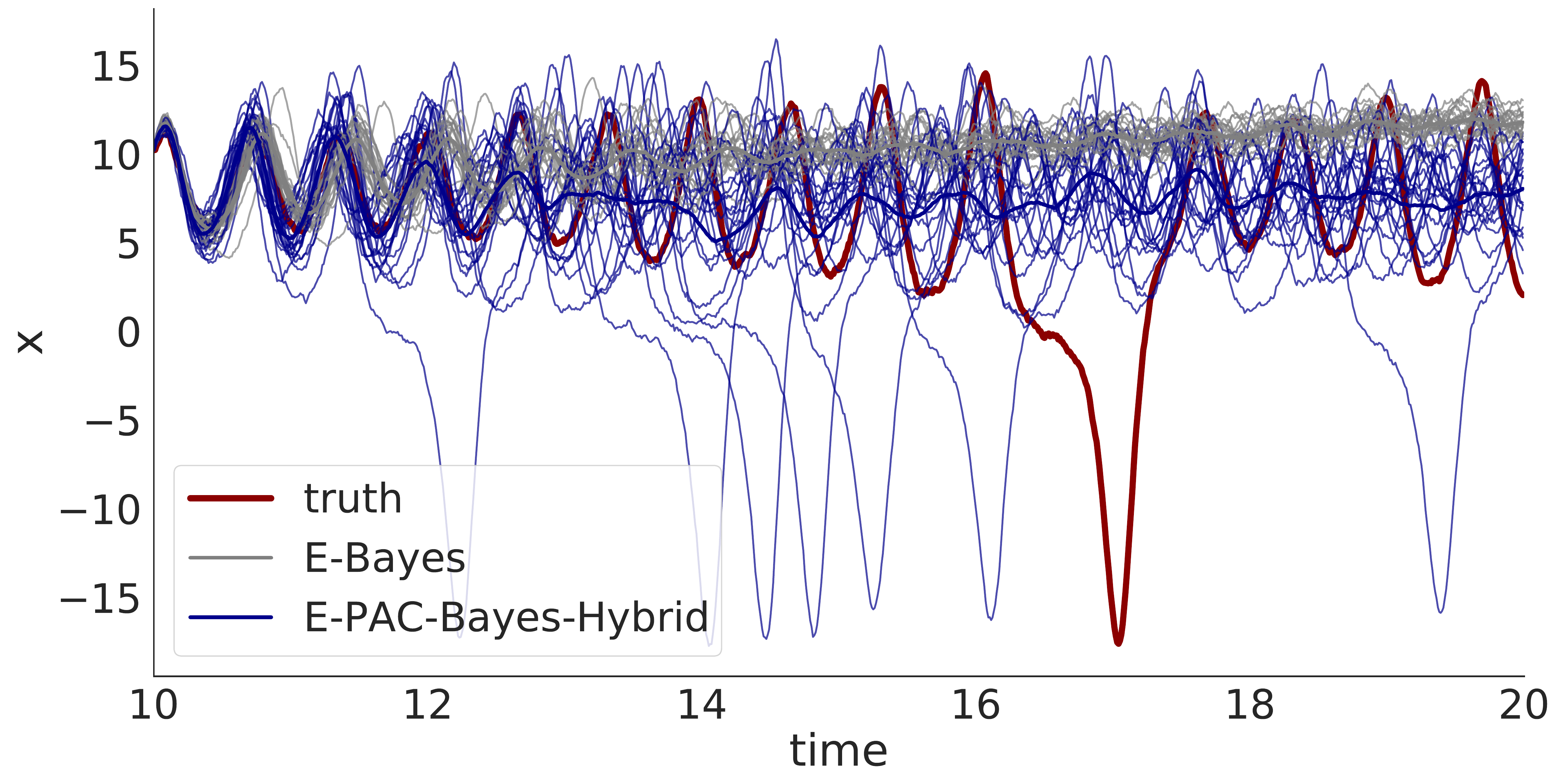}
\caption{$x$ coordinate over time}
\end{subfigure}
\begin{subfigure}{0.8\linewidth}
\centering 
\includegraphics[width=\linewidth]{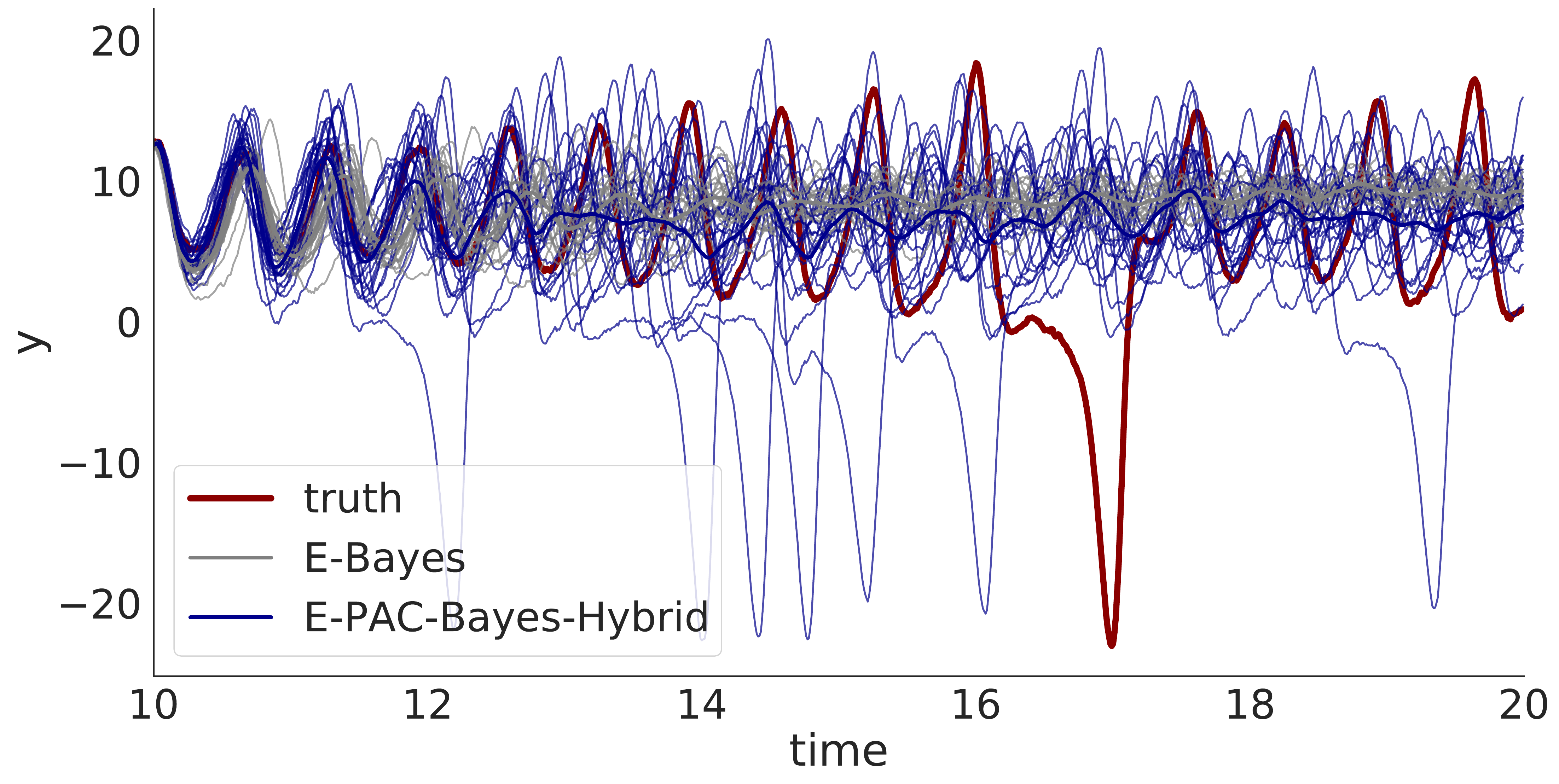}
\caption{$y$ coordinate over time}
\end{subfigure}
\begin{subfigure}{0.8\linewidth}
\centering 
\includegraphics[width=\linewidth]{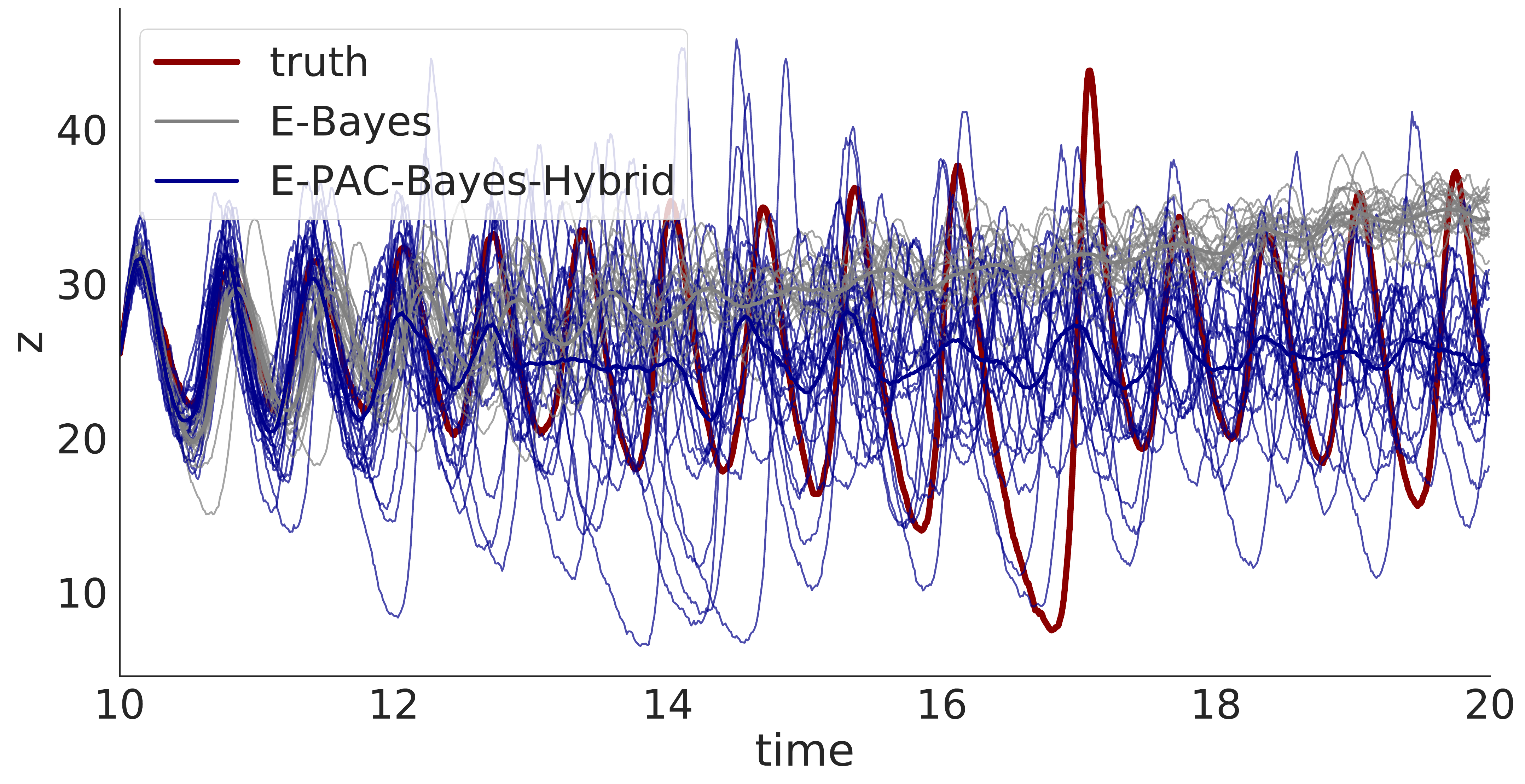}
\caption{$z$ coordinate over time}
\end{subfigure}
\caption{Predicting 1000 time steps ahead. Shows individual trajectories.}
\label{fig:lorenz1000raw}
\end{figure}

In order to properly visualize it we switch from the 3d plot to 1d plots showing always one of the three dimensions vs the time component. We always start at $T=10$, forcasting either 100 steps (as in the numerical evaluation), 200 or 1000 steps. All the following figures show the mean trajectory averaged over 21 trajectories, as well as an envelope of $\pm$ 2 standard deviations.
Figure~\ref{fig:lorenz100} visualizes that at that time scale the qualitative behavior is similar without clear differences. Doubling the predicted time interval as shown in Figure~\ref{fig:lorenz200} the baseline starts to diverge from the true test sequence, while our proposed model still tracks it closely be it at an increased variance. Finally predicting for 1000 time steps (Figure~\ref{fig:lorenz1000}) the chaotic behavior of the Lorenz attractor becomes visible as the mean in both setups no longer tracks the true trajectory. Note however that the baseline keeps has rather small variance and a strong tendency in its predictions that do not replicate the qualitative behavior of the Lorenz attractor. While the proposed model also shows an unreliable average, the large variance, which nearly always includes the true trajectory shows that the qualitative behavior is still replicated properly by individual trajectories of the model. See Figure~\ref{fig:lorenz1000raw} for seven individual trajectories of each of the two models. All trajectories of \emph{E-PAC-Bayes-Hybrid} show the qualitatively correct behavior, including even the characteristic loop. 

\end{document}